\documentclass{article}
\ifx\XeTeXversion\undefined\pdfoutput=1\fi

\usepackage{iclr2027_conference,times}
\iclrfinalcopy

\usepackage[utf8]{inputenc}
\usepackage[T1]{fontenc}
\PassOptionsToPackage{hyphens}{url}
\usepackage{hyperref}
\hypersetup{hidelinks,pdfborder={0 0 0}}
\usepackage{url}
\usepackage{booktabs}
\usepackage{longtable}
\usepackage{multirow}
\usepackage{siunitx}
\usepackage{amsfonts}
\usepackage{amsmath, amssymb}
\usepackage{nicefrac}
\usepackage{microtype}
\usepackage{xcolor}
\usepackage{graphicx}
\usepackage{tabularx}
\usepackage[section]{placeins}
\graphicspath{{figures/}}

\title{\raggedright Auditing Routing Entropy as an Uncertainty Signal in Attention-Residual Transformers}

\author{\begin{tabular}{c@{\hspace{2.5em}}c@{\hspace{2.5em}}c}
Wenhao Liang & Lin Yue & Wei Emma Zhang\\
Mingyu Guo & Olaf Maennel & Weitong Chen\\[4pt]
\multicolumn{3}{c}{\normalfont Adelaide University, Adelaide, Australia}
\end{tabular}}

\begin{document}
\maketitle

\begin{abstract}
Dynamic architectures leave a per-example routing trace beside each prediction,
and diffuse routing is easy to read as a sign that the prediction is
unreliable. We audit that reading for routing entropy in Attention-Residual
(AR) variants of Swin-Tiny and DeiT-Small, trained from scratch on CIFAR-10/100
with a soft-binned calibration auxiliary loss, asking whether the trace carries
information about correctness beyond what the model's own confidence already
reveals. Three checks probe this increment: does a routing signal appear at
fixed confidence, does it replicate across training seeds, and can a held-out
predictor exploit it against output-only and shuffled-trace controls? A
sensitivity audit then injects effects of known size and measures the fraction
of each that the probes recover. No test in the fixed 30-test binned family
survives multiplicity correction, and neither the nominal hit nor a borderline
result recurs in its sibling seeds. Across 24 paired runs a scalar routing
probe yields no pooled improvement in routing-stratified calibration, and an
entropy-profile probe predicts correctness better than the same probe given
shuffled profiles yet worse than a confidence-only predictor in both binary
log-loss and Brier score: a gain over shuffled traces does not become a gain
over the output. Conditioning on the complete logit vector leaves the
corresponding comparison unresolved. The audit bounds how far these
non-detections can be read: at an injected effect of 0.010 nats the profile
probe recovers 24--59\% of the oracle gain, and a reference-preserving
correction probe recovers 8\% and 23\% in the two CIFAR-100 settings, below
the threshold we fixed for applying it to real labels. The results establish
control-dependent gains and incomplete estimator recovery, not the absence of
conditional routing information.
\end{abstract}

\section{Introduction}
\label{sec:intro}

\begin{figure}[t]
  \centering
  \includegraphics[width=\linewidth]{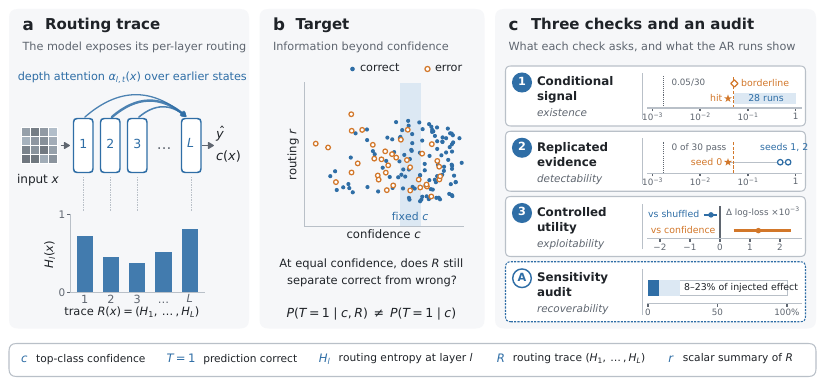}
  \caption{\textbf{From a routing trace to a testable uncertainty claim.}
  \emph{(a)} An AR vision transformer exposes top-class confidence $c(x)$ and,
  at every layer, routing weights over earlier states; their normalised entropies
  $H_l\in[0,1]$ form the trace $R(x)$ (input and values illustrative; the token
  index $n$ of Eq.~\eqref{eq:routing-entropy} is suppressed). \emph{(b)} The
  target: at equal confidence, does $R$ change the probability of a correct
  prediction? (schematic data). \emph{(c)} The three checks and the audit with the
  observed results: (1)~within-bin permutation $p$-values of the 30-test family,
  with one nominal hit and one borderline result at $p\approx0.05$ (\S\ref{sec:rq1});
  (2)~the hit is not reproduced by seeds 1--2 and no test passes $0.05/30$;
  (3)~Profile minus shuffled-profile and minus confidence-only binary log-loss
  with 95\% intervals ($\times10^{-3}$ nats; positive is worse for the profile;
  \S\ref{sec:probe}); (A)~recovery of a $0.010$-nat injected effect by the
  reference-preserving correction probe on the two CIFAR-100 settings
  (\S\ref{sec:richer-main}).}
  \label{fig:overview}
\end{figure}

Many networks choose their computation per input. Mixture-of-experts layers
route tokens to experts, adaptive-depth models skip layers, and token-pruning
systems decide which tokens to keep~\citep{shazeer2017outrageously,fedus2022switch,wang2018skipnet,rao2021dynamicvit}.
Each choice leaves a per-example trace next to the prediction, and it is natural
to ask whether the trace also signals when that prediction is unreliable.

A trace can be associated with errors and still carry nothing the prediction
does not already reveal. Routing may follow input difficulty, and difficulty
usually shows up in the model's confidence. The useful question is therefore
incremental. Let $T=\mathbf{1}[\hat y=y]$ indicate a correct prediction, $c$ the
top-class confidence and $R$ a routing representation. We ask whether
\begin{equation}
P(T=1\mid c,R)\neq P(T=1\mid c)
\label{eq:conditional-target}
\end{equation}
on a set of $(c,R)$ values with positive probability: among predictions of equal
confidence, does the route separate correct predictions from errors? The primary
analyses condition on $c$, as a post-hoc calibrator would. A secondary analysis
conditions on the complete logit vector. The two are different questions rather
than a weaker and a stricter version of one: conditional mutual information is
not monotone in the conditioning set, so $R$ can be uninformative given $c$ and
informative given the logits, or the reverse (App.~\ref{app:richer-output}).

Attention Residuals~\citep[AR;][]{team2026attention} make the question concrete.
AR replaces fixed residual accumulation with input-dependent attention over
earlier representations, so every forward pass exposes a normalised routing
distribution at each layer. Diffuse routing is easy to read as uncertainty, but
it may equally reflect a useful mixture of representations. The training
objective does not ask routing entropy to predict correctness, so whether it
does is a property of the trained models to be measured, not a consequence of
the architecture.

We measure it with three checks and a sensitivity audit (Fig.~\ref{fig:overview}c).
The \emph{conditional-signal} check compares correctness across routing-defined
groups at fixed confidence; \emph{replicated evidence} repeats that test across
training seeds within a fixed multiplicity family; \emph{controlled utility} asks
whether a held-out predictor gains from routing relative to output-only and
shuffled-trace controls. Each check addresses a different misreading of a feature
ablation: coarse confidence bins can leave residual confounding, a favourable
seed can overstate stability, and a flexible probe can profit from an extra input
that says nothing new about correctness. The \emph{sensitivity audit} addresses the
opposite misreading, that a failed probe demonstrates absence: it injects effects
of known size into simulated labels and measures how much of them each probe
recovers.

The choice of control changes what the routing probe appears to show. Across 24
paired runs, an entropy-profile probe attains lower binary log-loss and Brier
score than the same probe given shuffled profiles, but higher loss than a
confidence-only predictor. A gain over shuffled traces therefore does not
establish predictive value beyond the output. Replication plays the same role
for the binned diagnostic: a nominal hit and a borderline result are not
reproduced by their sibling seeds, and no test survives correction. The audit, in
turn, limits how these non-detections can be read. At an injected effect of 0.010 nats
per example the profile probe recovers between a quarter and three fifths of the
oracle gain, and a probe that keeps the output-only reference fixed recovers
less.

We make three contributions. (i) A replicated test of AR routing entropy at
fixed confidence, with a fixed 30-test family and continuous-confidence checks
on all 24 replicated runs, in which no apparent routing effect replicates.
(ii) Paired held-out comparisons in which the profile probe's advantage over
shuffled traces does not survive an output-only baseline conditioned on
confidence, while the analogous comparison against the complete logits remains
unresolved. (iii) A sensitivity audit showing that the fitted probes recover only
part of known routing effects and can lose to their own baseline when such an
effect is present, which limits how far a non-detection can be read. Conditional
probing and oracle proper-score identities are established
tools~\citep{hewitt2021conditional,xu2020theory}; what is new is the evidence
they yield about an uncertainty claim for internal routing.

\section{Routing Information Beyond Confidence}
\label{sec:method}

\subsection{Target and routing representation}
\label{sec:method-arcondcal}

Write $\eta_{c,R}=P(T=1\mid c,R)$ and $\eta_c=P(T=1\mid c)$. An oracle that
observes $(c,R)$ improves on one that observes only $c$ by
$\mathbb{E}[(\eta_{c,R}-\eta_c)^2]$ in binary Brier risk and by $I(T;R\mid c)$
in binary log-loss (App.~\ref{app:theory}). Fitted predictors fall short of their
oracles. If $f_c$ and $f_{c,R}$ have excess log-loss risks $e_c$ and $e_{c,R}$
above their respective oracles on the same evaluation distribution, then
\begin{equation}
L(f_c)-L(f_{c,R}) \;=\; I(T;R\mid c) + e_c - e_{c,R}.
\label{eq:excess-risk}
\end{equation}
We therefore use injected-signal audits to assess how much of a known effect the
principal probes recover: a fitted contrast can be negative even when
$I(T;R\mid c)>0$, if routing makes the predictor harder to fit.

At AR layer $l$, learned weights $\alpha_{l,t,n}(x)$ form a distribution over the
$T_l$ earlier states $t$ available to token $n$. In our implementation, Full AR
keeps every earlier state of the current stage (one per layer in Swin-Tiny, one
per sub-layer in DeiT-Small), whereas Block AR keeps the stage input and one
state per block of $b$ transformer layers ($b=2$ for Swin-Tiny and $b=1$ for
DeiT-Small; App.~\ref{app:details}). We summarise concentration by normalised
Shannon entropy averaged over the $N$ tokens,
\begin{equation}
H_l(x)=\frac{1}{N\log T_l}\sum_{n=1}^{N}
\left(-\sum_{t=1}^{T_l}\alpha_{l,t,n}(x)\log\alpha_{l,t,n}(x)\right)\in[0,1],
\label{eq:routing-entropy}
\end{equation}
omitting layers with $T_l=1$. The routing-entropy profile is
$R=(H_1,\ldots,H_L)$, with $L=8$, $16$, $22$ and $23$ for Swin Block, Swin Full,
DeiT Block and DeiT Full. Two scalar projections serve different roles: aggregate
entropy $r_{\rm agg}=L^{-1}\sum_lH_l$ defines routing-concentration groups for the
diagnostic, and depth variation $r_{\rm std}=\operatorname{std}_lH_l$ is the
feature of the scalar calibration probe. The profile keeps every $H_l$ but still
summarises, rather than retains, the raw routing weights; other scalar summaries
are described in App.~\ref{app:mechanism}.

\subsection{Three checks and a sensitivity audit}
\label{sec:falsification-ladder}

\emph{Conditional signal.} For each run, the diagnostic splits the 10,000 test
examples into $r_{\rm agg}$ tertiles, bins confidence into 15 equal-width bins,
and takes the maximum absolute accuracy gap between the low and high tertiles
over bins in which both groups have at least five examples. Its $p$-value is
$(1+\#\{\text{null}\geq\text{observed}\})/(B+1)$ over $B=5{,}000$ permutations of
$r_{\rm agg}$ within confidence bins. Because bins only approximate equal
confidence, a continuous-confidence check adds cross-fitted logistic correctness
models on $\operatorname{logit}(c)$ with and without routing, and permutation
tests within local confidence windows (App.~\ref{app:gap-robustness}).

\emph{Replicated evidence.} Each replicated architecture--dataset cell has three
training seeds. The multiplicity family is fixed at 30 tests, the 24 replicated
runs and six single-seed ViT-B/16 runs, with Bonferroni threshold $0.05/30$.

\emph{Controlled utility.} Each routing probe is compared with an output-only
predictor from the same model family on the same splits. Following
\citet{hewitt2021conditional}, the output-only arm receives zero padding in place
of routing inputs, which matches nominal parameterisation but not effective
capacity. The profile analysis adds a shuffled-trace control that keeps the
learner and input dimension but reassigns whole profiles across examples. The
shuffle preserves $P(R)$ but not $P(R\mid c)$; because $R$ and $c$ are
correlated, a profile arm can beat its shuffled arm without carrying information
beyond $c$.

\emph{Sensitivity audit.} We simulate binary labels $T^*$ under specified laws
$Q$ built from each run's observed confidences and profiles. $Q$ draws a
confidence anchor uniformly and a whole profile from the anchor's confidence
block; this preserves both empirical marginals but smooths their joint
distribution, so $Q$ is an alternative we specify, not a copy of the real
data-generating law. A fixed routing effect enters on the logit scale while
$\mathbb{E}_Q[T^*\mid c]$ matches a confidence-only fit, and its strength is set
so that the oracle binary log-loss gain $I_Q(T^*;R\mid c)$ equals 0, 0.0005,
0.002, 0.005 or 0.010 nats per example. Probes are fitted to sampled labels and
scored against the known label probabilities (expected-label loss). Recovery is
a probe's fitted gain divided by the oracle gain; it measures captured predictive
gain, not the rejection rate of a test, which we report separately for the
diagnostic. For the profile probe, a screen fixed before the audit required
positive Monte Carlo lower bounds at 0.005 and 0.010 nats and at least 25\%
recovery at 0.010 before weaker strengths or larger simulations were run
(App.~\ref{app:profile-sensitivity}); the correction probe of \S\ref{sec:richer-main}
has its own continuation rule.

\subsection{Probes}
\label{sec:ar-condcal}

\emph{Scalar probe.} AR-CondCal serves as a controlled scalar probe, not a
proposed calibrator: it tests whether $r_{\rm std}$ is exploitable within one
kernel family. From calibration triples $(c_i,r_i,T_i)$ with $r=r_{\rm std}$, it
estimates $P(T=1\mid c,r)$ by two-dimensional Gaussian Nadaraya--Watson
regression~\citep{nadaraya1964estimating,watson1964smooth},
\begin{equation}
\hat g(c,r)=\frac{\sum_iK_h(c-c_i,r-r_i)T_i}{\sum_iK_h(c-c_i,r-r_i)}.
\label{eq:scalar-probe}
\end{equation}
Its control NW($c$) is one-dimensional; both use the Gaussian kernel and Scott's
rule for their own dimension. Each estimate is projected into
$[1/K+\epsilon,\,1-\epsilon]$ and mapped back to class probabilities by the
per-example temperature whose top-class probability equals the projected value,
which preserves Top-1 predictions (App.~\ref{app:impl}).
\label{sec:stress-metric}
The routing-defined stress score is \emph{worst-tertile ECE}: split the
evaluation examples into three equal-count $r_{\rm std}$ groups, compute 10-bin
equal-width ECE within each, and keep the largest (App.~\ref{app:wt-ece-illustration}).
Both methods are scored on the same groups.

\emph{Profile probe.} The profile arm predicts $T$ from $(c,H_1,\ldots,H_L)$; the
confidence-only and shuffled arms replace the profile by zeros or by shuffled
profiles. Each arm chooses, by inner-validation binary log-loss, between a
one-hidden-layer MLP with 32 units and a shared ridge-logistic model of
$\operatorname{logit}(c)$, and the same criterion selects the MLP's training
epochs. Outer five-fold cross-fitting is repeated twice; all preprocessing,
selection and shuffling happen within training partitions
(App.~\ref{app:probe-sanity}).

\pagebreak[2]
\section{Experimental Setup}
\label{sec:experiments}

The replicated design has 24 runs: two backbones
\citep[Swin-Tiny and DeiT-Small;][]{liu2021swin,touvron2021training} $\times$ two
datasets (CIFAR-10/100) $\times$ two AR variants (Block, Full) $\times$ three
training seeds. Six single-seed ViT-B/16 runs~\citep{dosovitskiy2020image} enter
the diagnostic family but no replicated or held-out analysis. Models are trained
from scratch with the DeiT recipe (mixup, CutMix, label smoothing) plus a
soft-binned ECE auxiliary loss~\citep[weight 0.1;][]{karandikar2021soft} and are
under-confident before calibration (App.~\ref{app:reliability}). According to the
training logs, 21 runs follow one uninterrupted 300-epoch schedule; three CIFAR-10
seed-0 runs deviate in batch size or schedule, so these replicates differ in more
than the seed (App.~\ref{app:details}). Test accuracy ranges from 0.68 to 0.76 on
CIFAR-100 and from 0.69 to 0.92 on CIFAR-10 (App.~\ref{app:provenance}). Two
lower-accuracy CIFAR-10 seed-0 runs (0.69 and 0.74), both among the three, are
retained under the fixed inclusion rule; post-hoc checks omit them.

The diagnostic uses all 10,000 test examples and fits nothing. Scalar calibrators
use a fixed 5,000/5,000 calibration/test split. The profile and complete-output
probes use cross-fitting over all 10,000 examples. Worst-tertile ECE is the
primary scalar score, accompanied by multiclass NLL and Brier score of the
calibrated class probabilities; the correctness probes use binary log-loss and
Brier score, with AUROC secondary. Real-data contrasts are reported as
$\Delta L=L(\text{routing arm})-L(\text{comparator})$, so negative values favour
routing; binary log-loss differences are in $10^{-3}$ nats per example and Brier
differences in $10^{-3}$. Simulation results are reported as gains
$G=L(\text{comparator})-L(\text{routing arm})$, positive favouring routing.

Pooled intervals come from a hierarchical bootstrap that resamples the four
backbone--dataset settings and then runs within them. Four clusters support only
exploratory, procedure-specific intervals: the 24 runs replicate within settings
and are not 24 independent settings. With six runs per setting, record-weighted
and equal-setting means coincide. Leave-one-setting-out means are reported for
the profile and complete-output analyses (App.~\ref{app:probe-sanity},
\ref{app:richer-output}).

\section{Results}

\subsection{Conditional signal and replication}
\label{sec:rq1}
\label{sec:rq2}

\begin{figure}[t]
  \centering
  \includegraphics[width=\linewidth]{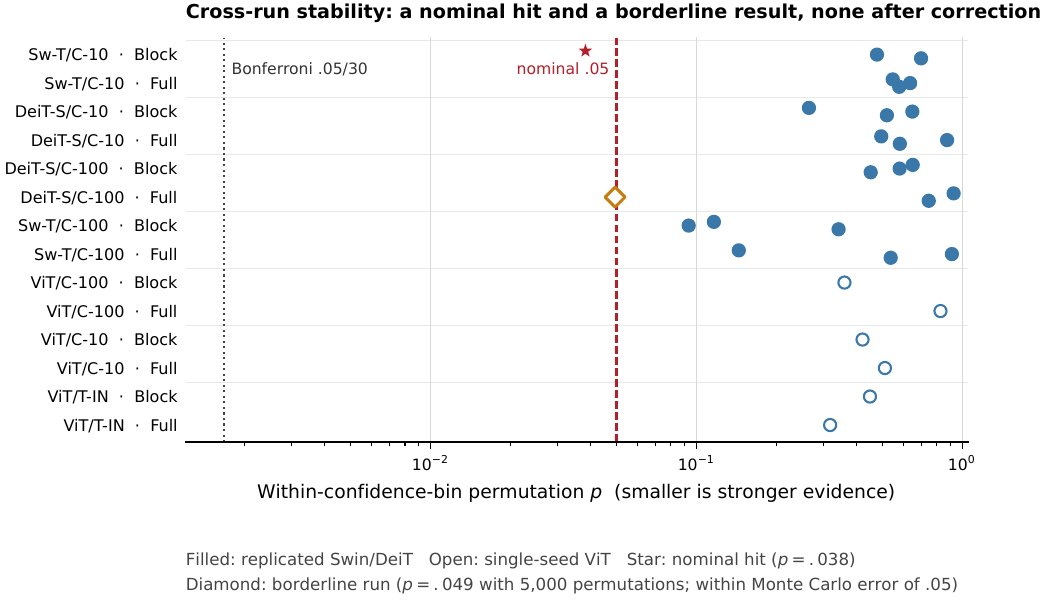}
  \caption{Within-bin permutation $p$-values (5,000 permutations) for the fixed
  30-test $r_{\rm agg}$ family. Rows are backbone--dataset/AR cells; filled points
  are replicated Swin/DeiT runs and open points single-seed ViT runs. The star marks
  the nominal hit and the diamond the borderline run, whose $p$ lies within Monte
  Carlo error of $0.05$; vertical lines mark $0.05$ and the Bonferroni threshold
  $0.05/30$.}
  \label{fig:cross-run-stability}
\end{figure}

Neither apparent routing effect in the binned diagnostic replicates. Swin-Tiny
Block-AR on CIFAR-10 seed 0 rejects nominally ($p=0.038$, maximum gap $0.345$),
but seeds 1 and 2 give $p=0.477$ and $0.699$ (gaps $0.132$ and $0.100$).
DeiT-Small Full-AR on CIFAR-100 seed 1 is borderline: $p=0.049$ with 5,000
permutations (Monte Carlo standard error $0.003$), and five recomputations with
50,000 permutations each range from $0.048$ to $0.050$; its sibling seeds give
$p=0.927$ and $0.747$ (App.~\ref{app:replication-plan}). No test passes
$0.05/30=0.0017$, and none would pass $0.05/24$ if the family were restricted to
the replicated runs (Fig.~\ref{fig:cross-run-stability}). Uniform null
$p$-values would give 1.5 nominal rejections in expectation; this is a
reference point, not evidence about the joint null. Nor is the absence of a
Bonferroni rejection strong evidence against moderate effects: with an injected
$r_{\rm agg}$ effect of 0.010 nats, the per-run rejection rate at $0.05/30$
ranges from $0.185$ to $0.85$ across the four settings
(Table~\ref{tab:scalar-sensitivity}).

Confidence bins only approximate conditioning on $c$. We therefore extended the
continuous-confidence check, originally run on four representative runs, to all
24 replicated runs after the borderline result (App.~\ref{app:gap-robustness}).
It finds no consistent routing improvement. Adding $r_{\rm agg}$ to a cross-fitted
confidence-only logistic model never improves held-out log-loss with an interval
excluding zero, and three runs get worse. Adding the whole profile improves one
run (Swin-Tiny Full-AR, CIFAR-100, seed 1: $1.7\times10^{-3}$ nats, interval
$[0.3,3.1]\times10^{-3}$), an uncorrected improvement that its sibling seeds do
not repeat, and worsens one. Local-window permutation tests, which we treat as
descriptive, give $p<0.05$ in 5 of 96 run--window combinations (4.8 if every null
held; the count is not evidence for the joint null): three windows of one
DeiT-Small Block-AR CIFAR-100 run whose siblings show no association, one of the
borderline run, and one other.

\subsection{Scalar routing and calibration}
\label{sec:rq3}

\begin{figure}[t]
  \centering
  \includegraphics[width=\linewidth]{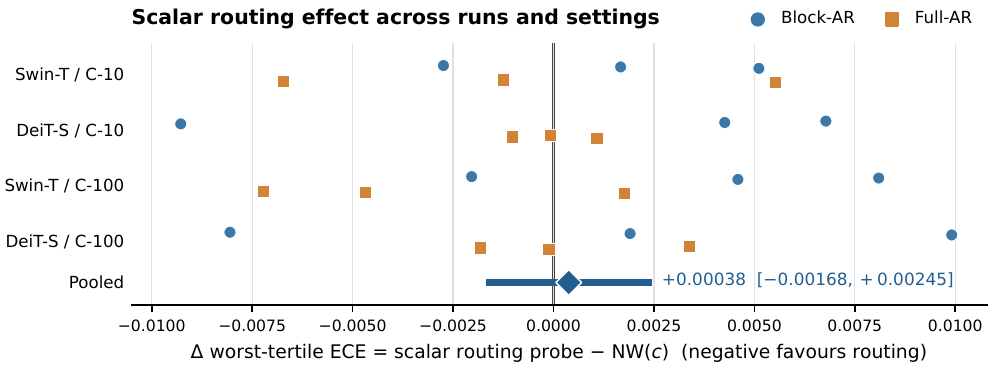}
  \caption{Scalar probe minus NW($c$) in worst-tertile ECE for each of the 24 runs,
  grouped by backbone--dataset setting (circles Block-AR, squares Full-AR). The
  pooled diamond and bar show the mean and the setting-clustered 95\% interval; no
  setting-specific interval is inferred.}
  \label{fig:scalar-effect}
\end{figure}

Adding $r_{\rm std}$ does not improve the pooled routing-defined stress score.
Across 24 paired runs, the scalar probe minus NW($c$) has pooled worst-tertile ECE
$+0.00038$ (95\% interval $[-0.00168,+0.00245]$; Fig.~\ref{fig:scalar-effect});
12 of the 24 differences are negative, and they range from $-0.0093$ to $+0.0099$.
Under this four-setting bootstrap, pooled improvements larger than about $0.0017$
lie outside the interval (App.~\ref{app:effect-size}). The calibrated class
probabilities become worse with routing in multiclass NLL, $+0.0084$
$[+0.0061,+0.0109]$, and in Brier score, $+0.0010$ $[+0.0008,+0.0013]$.

The simulations illustrate a possible source of this estimator cost: under the
specified null laws, the two-dimensional estimator has expected binary log-loss 3.4
to $5.2\times10^{-3}$ nats higher than NW($c$), and it is still worse at injected
effects of 0.0005 and 0.002 nats; it recovers 54--65\% of a 0.010-nat effect
(Table~\ref{tab:scalar-sensitivity}). By Eq.~\eqref{eq:excess-risk}, under these
laws a scalar routing effect of the smaller sizes would be absorbed by the
estimator's extra excess risk rather than appear as a gain; the simulations do not
isolate the cause of the real-data cost. The comparison also leaves open whether
compressing the profile to one number discards useful layer-wise information, so
the next comparison keeps the entire profile.

\subsection{The entropy profile: two controls, two answers}
\label{sec:probe}

\begin{figure}[t]
\centering
\includegraphics[width=0.92\linewidth]{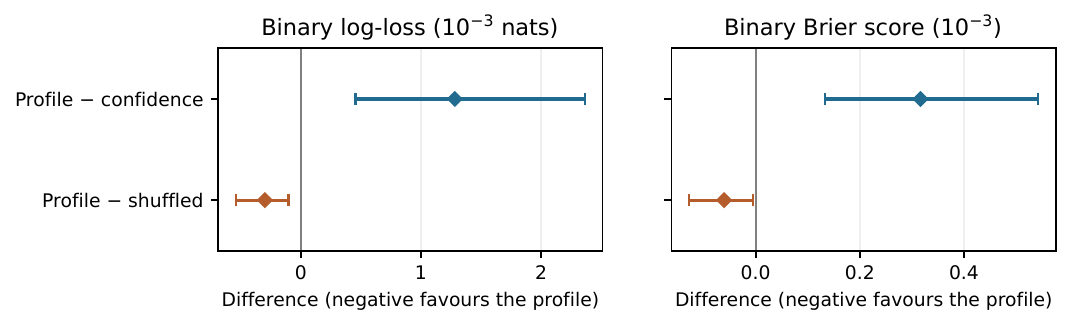}
\caption{\textbf{Entropy-profile comparisons across 24 paired runs.} Profile
minus comparator in binary log-loss ($10^{-3}$ nats) and Brier score ($10^{-3}$);
negative values favour the profile. Points are pooled means (record-weighted and
equal-setting means coincide) with 95\% hierarchical-bootstrap intervals over the
four settings. The two comparators support different conclusions.}
\label{fig:profile-control-chain}
\end{figure}

The profile probe beats shuffled profiles but not the confidence-only predictor
(Fig.~\ref{fig:profile-control-chain}). Pooled over the 24 runs, profile minus
confidence-only is $+1.28$ $[+0.45,+2.36]$ in binary log-loss and $+0.32$
$[+0.13,+0.54]$ in Brier score, whereas profile minus shuffled profiles is
$-0.30$ $[-0.54,-0.10]$ and $-0.061$ $[-0.128,-0.005]$ (units as in
\S\ref{sec:experiments}; Table~\ref{tab:repaired-primary}). Both orderings hold
in every setting mean and every leave-one-setting-out mean. A study that used
only the shuffle control would conclude that the profile carries predictive
information; the output-only comparison shows that the same probe predicts
correctness worse than confidence alone. Because the shuffle also breaks the
dependence between $R$ and $c$, it cannot serve as a conditional null.
Post-hoc checks that omit the runs whose training deviated from the standard
recipe preserve both orderings in log-loss; the Brier interval against shuffled
profiles reaches or crosses zero whenever the two lower-accuracy runs are omitted
(App.~\ref{app:probe-sanity}).

Recoverability limits how the confidence-only comparison can be read. In the
audit the profile probe recovers 24--59\% of a 0.010-nat injected effect across
the four settings, but 6--32\% at 0.005 nats and none on DeiT-Small/CIFAR-100
($-11\%$), which failed the audit's screen and stopped further simulation
(App.~\ref{app:profile-sensitivity}). Failing to improve on confidence therefore
does not bound $I(T;R\mid c)$ tightly: in Eq.~\eqref{eq:excess-risk}, a positive
information term can be offset by the larger excess risk of the
routing-augmented learner.

\subsection{Conditioning on the complete output}
\label{sec:richer-main}

Confidence discards most of the output, so the profile could fail against $c$
and still add information beyond the full logits. The complete-output comparison
conditions on centred logits, their mean, $\operatorname{logit}(c)$, predictive
entropy and the top-two margin, a lossless recoding of the logits plus three
summaries (App.~\ref{app:richer-output}). Each arm again selects an MLP or falls
back to a shared ridge-logistic reference fitted on the output alone. Under this
selection rule profile minus output-only is $+0.87$ $[-0.05,+2.46]$; with
ridge-logistic learners in all arms it is $+0.21$ $[-0.03,+0.46]$. Both intervals
contain zero, so the ordering seen under confidence conditioning is preserved but
not established. Against permuted profiles the profile arm is better with either
learner (MLP $-0.88$ $[-1.52,-0.23]$; ridge $-0.30$ $[-0.58,-0.03]$). On the
twelve CIFAR-100 runs ($K=100$ classes) all 360 fold-level selections
(12 runs $\times$ 10 folds $\times$ 3 arms) returned the reference, so the selected
contrasts there are exactly zero; the unselected routing MLP trailed the
reference in every such run, by $7.6\times10^{-3}$ nats on average. 
Because the largest weight decay of the MLP grid was chosen in 462 of 720
selections, we refitted all 24 runs post hoc with the grid extended from
$\{10^{-4},10^{-3},10^{-2}\}$ to include $10^{-1}$ and $1$ for every MLP arm, changing nothing else.
The selected contrast becomes $+0.35$ $[-0.18,+1.20]$, the MLP arms keep their
orderings (output-only $+0.75$ $[+0.25,+1.46]$; permuted $-0.78$
$[-1.19,-0.39]$) and fallbacks drop from 563 to 407; the new maximum is still
chosen in 636 selections, so the adequacy of these learners at the edge of the
grid remains open (App.~\ref{app:richer-output}).

The injected-signal control of App.~\ref{app:richer-output} shows this
reversion occurring with a routing effect present: the routing-augmented
network there beats its output-only counterpart in $20/20$ replicates in seven
of the eight positive-strength cells. At $0.010$ nats, however, the gain closes
only $21.8\%$ and $57.4\%$ of the output-only network's deficit against the
reference on Swin/CIFAR-100 and DeiT/CIFAR-100, respectively, leaving the
augmented candidate behind. A follow-up asks whether the real-data
reversion likewise reflects a weak output-only learner rather than missing
information. Designed after these results and run on the same
injected-signal simulation records, so not an independent confirmation, the
correction probe keeps the ridge reference fixed and learns only a log-odds
correction, with the same budget for an output-only correction. Its routing
increment has a positive 95\% Monte Carlo lower bound in all four
positive-strength conditions (Table~\ref{tab:residual-probe}). At 0.010 nats it
recovers 8.1\% of the oracle gain on Swin-Tiny/CIFAR-100 and 22.6\% on
DeiT-Small/CIFAR-100, a mean of 15.4\%, below the 25\% continuation threshold
fixed before the follow-up ran. We therefore did not evaluate the probe on real
correctness labels, and its behaviour there is unknown. Keeping the reference
also moves the candidate from 2.7--$7.9\times10^{-3}$ nats behind it to level or
ahead; that movement mixes baseline improvement with routing and is not
recovered routing information.

% Generated by scripts/e3_residual_probe/make_table_e3.py from
% audit_sources/e3_residual_probe/corrected_quantities.csv and PHASE1_GATE.json.
\begin{table}[t]
\centering\small
\caption{\textbf{Injected-signal simulations for the complete-output probe (E2) and
the reference-preserving correction probe (E3)} on the two CIFAR-100 settings, 20
replicates per condition. Unlike Fig.~\ref{fig:profile-control-chain} and
Tables~\ref{tab:repaired-primary} and~\ref{tab:e2-richer-output}, entries are
\emph{gains} in expected-label binary log-loss ($10^{-3}$ nats per example), so
positive values favour the routing arm. The routing increment
$G_{\rm route}=L(A_{\rm sel})-L(B_{\rm sel})$ compares the selected output-only and
routing arms under the same budget and is the quantity that measures routing (95\%
Monte Carlo intervals for E3); recovery divides the E3 increment by the exact oracle
gain. The candidate gain $G_{\rm cand}=L(\mathrm{REF})-L(B_{\rm raw})$ compares the
unselected routing candidate with the output-only ridge reference; it includes
baseline predictive improvement and is not recovered routing information. All
labels are simulated; on the real CIFAR-100 labels every E2 selection returned the
reference (\S\ref{sec:richer-main}).}
\label{tab:residual-probe}
\setlength{\tabcolsep}{3.5pt}
\begin{tabular}{llrrrrr}
\toprule
& Oracle & \multicolumn{2}{c}{Routing increment $G_{\rm route}$} & \multicolumn{2}{c}{Candidate gain $G_{\rm cand}$} & Recovery \\
\cmidrule(lr){3-4}\cmidrule(lr){5-6}
Setting & gain & E2 & E3 & E2 & E3 & (E3) \\
\midrule
Swin / C-100 & 0 & $+0.00$ & $-0.02$~$[-0.06,+0.02]$ & $-7.86$ & $-0.09$ & -- \\
Swin / C-100 & 0.005 & $-0.02$ & $+0.28$~$[+0.21,+0.34]$ & $-6.67$ & $+0.19$ & 5.5\% \\
Swin / C-100 & 0.010 & $-0.02$ & $+0.81$~$[+0.72,+0.91]$ & $-5.71$ & $+0.74$ & 8.1\% \\
\midrule
DeiT / C-100 & 0 & $+0.06$ & $-0.02$~$[-0.06,+0.01]$ & $-7.71$ & $-0.05$ & -- \\
DeiT / C-100 & 0.005 & $-0.04$ & $+0.77$~$[+0.66,+0.89]$ & $-5.07$ & $+0.72$ & 15.5\% \\
DeiT / C-100 & 0.010 & $+0.07$ & $+2.26$~$[+2.11,+2.41]$ & $-2.66$ & $+2.19$ & 22.6\% \\
\bottomrule
\end{tabular}
\end{table}

\subsection{Robustness}
\label{sec:robustness}

Three further checks address artefacts of the analysis. Null simulations on the
observed $(c,r)$ support with 10--50 confidence bins show no detectable
anti-conservative behaviour of the binned test (largest Type-I estimate $0.0545$
at nominal $0.05$; App.~\ref{app:gap-robustness}). An eight-run bandwidth
analysis finds no bandwidth rule that consistently improves the scalar probe's
worst-tertile ECE (App.~\ref{app:bw-sensitivity}). On the tested Swin-Tiny
CIFAR-10 Block-AR run, the mean ECE of temperature scaling, MEC (a logits-only
calibrator; App.~\ref{app:internal-baselines}) and the scalar probe changes by less than
0.001 between 2,500 and 5,000 calibration examples (App.~\ref{app:calsize}). Metric
choice still matters, because equal-width ECE and SmoothECE rank calibrators
differently (App.~\ref{app:cwece}); we therefore read the stress score together
with proper scores.

\section{Related Work}
\label{sec:related}

Post-hoc calibration maps model outputs to probabilities
\citep{guo2017calibration,naeini2015obtaining}. Failure prediction can also draw
on hidden features: Trust Score compares a test point with training points in
representation space~\citep{jiang2018trust}, and ConfidNet learns true-class
probability from a classifier's representations~\citep{corbiere2019addressing}.
Probes on the hidden states of language models predict whether a statement is
true~\citep{azaria2023internal} or the semantic entropy of generated
answers~\citep{kossen2024semantic}. Our question is narrower: whether a routing
trace adds information about correctness once confidence is known.

Probe design decides what such an increment means. \citet{hewitt2019designing}
separate information in a representation from a probe's ability to learn the task
with control tasks. \citet{hewitt2021conditional} measure information beyond a
baseline by comparing a probe on $[\text{baseline};\text{representation}]$ with
the same probe on $[\text{baseline};\text{zero padding}]$, building on the
predictive $\mathcal{V}$-information of \citet{xu2020theory}, which makes usable
information depend on the predictor family. Our output-only arm is that
zero-padded construction with the model's own output as the baseline and
correctness as the target; the audit then asks how much of a known increment the
chosen family actually captures. Conditional permutation tests face a related
difficulty, since their guarantees depend on the conditional distribution used to
generate reference samples~\citep{berrett2020conditional}, and \citet{shah2020hardness}
show that, over the unrestricted class of absolutely continuous distributions, a test
whose level holds uniformly over the conditional-independence null has power no
greater than its level against any alternative; neither result removes the need to measure
a particular procedure's sensitivity. Our global shuffle is accordingly a
trace-destruction comparison, and the continuous-confidence analyses address the
conditional question.

Subgroup calibration motivates examining routing-defined
groups~\citep{hebert2018multicalibration}; our tertile score tests one specified
partition rather than a class of subgroups. ECE depends on binning and estimation
bias~\citep{kumar2019verified,nixon2019measuring,roelofs2022mitigating}, and
SmoothECE offers a kernel alternative~\citep{blasiok2023smooth}. Dynamic
computation supplies the candidate traces
\citep{shazeer2017outrageously,fedus2022switch,wang2018skipnet,rao2021dynamicvit}, and
AR extends input-dependent selection to residual
aggregation~\citep{team2026attention}; these works concern efficiency and
accuracy and make no claim about conditional uncertainty.

\section{Discussion}
\label{sec:discussion}

Our clearest result is a control comparison, not a claim of absence. The profile
probe improves on shuffled traces but not on confidence-only prediction, and the
audit shows why the second statement cannot be turned into ``routing carries no
information'': when the simulation law contains a known routing effect, the
fitted probes recover only part of it and can lose to their own baseline. Diffuse
routing may reflect a mixture of useful representations, or difficulty that
confidence already expresses; our results do not identify which applies to AR,
nor separate absent information from incomplete recovery.

The appropriate comparator depends on the claim. An increment beyond confidence
must be tested against a confidence-only predictor; an increment beyond the
complete output requires a predictor conditioned on the logits. The two claims
are not nested, because conditional mutual information need not fall when the
conditioning set grows. For studies of internal traces as uncertainty signals we
suggest three practices: compare with an output-only predictor from the same
family, not only with shuffled traces; replicate across training seeds within a
fixed multiplicity family; and report how much of an injected signal of
comparable strength the probe recovers before interpreting a null. The same
practices apply to expert assignments, token-retention patterns and
adaptive-depth decisions, although we have not tested those traces.

The evidence is limited in scope. The models are small, trained from scratch on
CIFAR-10/100 with a calibration auxiliary loss, and under-confident before
calibration; only four replicated backbone--dataset settings enter the pooled
analyses, three of the 24 runs deviate from the standard batch or schedule, and
ViT-B/16 contributes single-seed diagnostics only. We did not train without the
auxiliary loss, so the conclusions need not carry over to cross-entropy-only AR
models; nor can they be attributed to that loss, which does not force
$I(T;R\mid c)=0$. We test entropy summaries of routing, not the raw
routing weights, and the correction probe was not run on real correctness labels.
Whether the conclusions transfer to larger pretrained models, other routing
mechanisms, language models or distribution shift has not been studied.

\paragraph{Conclusion.}
In the Attention-Residual models studied here, an advantage over shuffled
entropy profiles does not become an advantage over confidence-only prediction,
and known-signal simulations reveal that the probes recover only part of a
routing effect when one exists. Evaluations of internal traces as uncertainty
signals should therefore report output-only comparisons, seed replication and
sensitivity to specified alternatives before drawing conclusions about added
information.
\label{sec:conclusion}

\noindent\textbf{Reproducibility statement.}
The 24 replicated runs are listed in Table~\ref{tab:run-accuracy} and the six
single-seed ViT-B/16 runs in Table~\ref{tab:vit-pilot-compact}. Training and
evaluation settings are in Apps.~\ref{app:details}--\ref{app:reproducibility}; the
probe protocols, including folds, seeds, selection rules and the injected-signal
generators, are in Apps.~\ref{app:gap-robustness}, \ref{app:probe-sanity}
and~\ref{app:residual-probe}. The supplementary archive contains the routing caches of
the 24 replicated runs, the code of every analysis, per-run results with a manifest
that maps each headline number to its source, a one-command verification from the
stored results, a cache-driven smoke test, and the training and extraction code with
the run configurations. Model checkpoints are not included (App.~\ref{app:provenance}).

% ===========================================================================
% The `thebibliography` environment emits its own "References" heading, so
% we do not add a separate \section*{References} (it would produce a
% duplicate heading).
\begingroup\small
\bibliographystyle{iclr2027_conference}
\bibliography{refs}
\endgroup

\newpage
% ---------------------------------------------------------------------------
% Appendix
% ---------------------------------------------------------------------------
\appendix

\section{Worst-tertile ECE: bin-level construction}
\label{app:wt-ece-illustration}

Worst-tertile ECE is the calibration stress score introduced in
\S\ref{sec:stress-metric}. It is distinct from the low--high accuracy
gap used by the binned diagnostic. The main text gives the intuition;
this appendix defines the cut points, tie convention, confidence bins,
and final aggregation.

\textbf{Step 1: tertile split by routing score.}
Given the evaluation set
$\mathcal{D}_{\mathrm{eval}} = \{(x_i, y_i)\}_{i=1}^{n}$ and a
per-sample routing feature $\rho_i = \rho(x_i)$, compute the
empirical $33$rd and $67$th percentiles $q_{1/3}, q_{2/3}$ of
$\{\rho_i\}_{i=1}^{n}$ and assign each sample to one of the index
sets $\mathcal{S}_{\text{low}}, \mathcal{S}_{\text{mid}},
\mathcal{S}_{\text{high}}$ as
\begin{equation}
\mathcal{S}_{\text{low}}=\{i:\rho_i\le q_{1/3}\},\quad
\mathcal{S}_{\text{mid}}=\{i:q_{1/3}<\rho_i\le q_{2/3}\},\quad
\mathcal{S}_{\text{high}}=\{i:\rho_i>q_{2/3}\}.
\label{eq:wt-tertiles}
\end{equation}
By construction each tertile contains approximately $n/3$ samples; ties use
the strict-inequality convention shown above.

\textbf{Step 2: equal-width confidence binning within each tertile.}
For tertile $t \in \{\text{low}, \text{mid}, \text{high}\}$ and
$b \in \{1, \dots, 10\}$, define the $b$-th equal-width confidence
bin restricted to that tertile,
\begin{equation}
\mathcal{R}_{t,b} \;=\; \bigl\{\, i \in \mathcal{S}_{t} \,:\,
   c(x_i) \in \bigl(\tfrac{b-1}{10},\; \tfrac{b}{10}\bigr] \,\bigr\},
\end{equation}
where $c(x) = \max_k p_k(x)$ is the top-class confidence used
throughout. The within-bin empirical accuracy and mean confidence are
\begin{equation}
\mathrm{acc}(\mathcal{R}_{t,b})
   \;=\; \frac{1}{|\mathcal{R}_{t,b}|}
        \sum_{i \in \mathcal{R}_{t,b}} \mathbf{1}[\hat y_i = y_i],
\qquad
\mathrm{conf}(\mathcal{R}_{t,b})
   \;=\; \frac{1}{|\mathcal{R}_{t,b}|}
        \sum_{i \in \mathcal{R}_{t,b}} c(x_i),
\end{equation}
with $\hat y_i = \arg\max_k p_k(x_i)$.

\textbf{Step 3: tertile-wise ECE and worst-tertile aggregation.}
The within-tertile equal-width $\mathrm{ECE}_{10}$ operator that
appears in the body definition expands as
\begin{equation}
\mathrm{ECE}_{10}(\mathcal{S}_{t})
   \;=\; \sum_{b=1}^{10}
        \frac{|\mathcal{R}_{t,b}|}{|\mathcal{S}_{t}|}\;
        \bigl|\,\mathrm{acc}(\mathcal{R}_{t,b})
              - \mathrm{conf}(\mathcal{R}_{t,b})\,\bigr|,
\end{equation}
i.e.\ equal-width ECE~\citep{guo2017calibration,naeini2015obtaining} with 10
bins computed on the sub-sample indexed by $\mathcal{S}_{t}$ (global ECE elsewhere
uses 15 bins). Bins are right-closed, so $c=1$ falls in the last bin, and a tertile
with fewer than 10 examples would be skipped (none is). Worst-tertile ECE is
\begin{equation}
\mathrm{ECE}^{\mathrm{wt}}(\mathcal{D}_{\mathrm{eval}})
=\max_{t\in\{\mathrm{low},\mathrm{mid},\mathrm{high}\}}
\mathrm{ECE}_{10}(\mathcal{S}_t).
\label{eq:wt-final}
\end{equation}
It reports the calibration error of the most miscalibrated routing-defined
subgroup rather than the global average over $\mathcal{D}_{\mathrm{eval}}$, so it
targets calibration heterogeneity along the feature used by the scalar probe. This
is a stress test of one specified partition; it does not favour either method by
construction.

\FloatBarrier
\section{Single-run diagnostic and descriptive calibration benchmark}
\label{app:benchmark-tables}

The two tables report the complete single-run diagnostic and 14-method
calibration benchmark. The cross-run diagnostic uses all 30 runs
(Fig.~\ref{fig:cross-run-stability}), while the scalar comparison uses the
paired 24-run analysis (Fig.~\ref{fig:scalar-effect}). The nominally positive
seed remains visible in both the table and the cross-run figure.

\textbf{Evaluation populations.} The diagnostic table uses the full
10,000-example test set because it fits no calibrator and benefits from larger
per-bin support. The calibration benchmark uses the fixed seed-42 5,000/5,000
calibration/test split; all methods are fit on the first half and evaluated on
the held-out half. Bootstrap intervals use 500 resamples of that test half.
Appendix~\ref{app:calsize} audits the calibration-set size.

\textbf{Benchmark catalogue.} Descriptive external baselines are Temperature,
Vector, and Classwise Scaling~\citep{guo2017calibration}; Ensemble
TS~\citep{zhang2020mix}; Parametric TS~\citep{tomani2022parameterized}; Histogram
Binning~\citep{zadrozny2001obtaining}; BBQ~\citep{naeini2015obtaining}; Isotonic
Regression~\citep{zadrozny2002transforming}; and SB-ECE
TS~\citep{karandikar2021soft}. MEC is the logits-only internal comparator
defined in App.~\ref{app:internal-baselines}. Temperature Scaling and Ensemble TS
agree to four decimals in Table~\ref{tab:calibration_block_ar} but are distinct
fits; their unrounded scores differ from the fifth decimal (for example, seed-0 ECE
$0.010540$ and $0.010532$). The final three rows form the
controlled NW family: NW($c$), NW($c,H_{\mathrm{pred}}$), and the scalar probe
NW($c,r_{\mathrm{std}}$).

\begin{table}[t]
\centering
\setlength{\tabcolsep}{3pt}
\footnotesize
\caption{\textbf{Single-run diagnostic on Swin-Tiny CIFAR-10 seed 0.}
Samples are split by $r_{\mathrm{agg}}$ tertile and compared within 15
confidence bins; Bins and Min/$q_{25}$/Med encode shared-bin support. Wt.\ CI
uses 5,000 bootstrap resamples, while Null $q_{.975}$ and $p$ use 5,000
within-bin routing permutations. This is the nominal hit of the 30-test family,
which also contains one borderline result; definitions, the borderline run and
caveats are in App.~\ref{app:gap-robustness}.}
\label{tab:phenomenon}
\begin{tabular}{lccccccc}
\toprule
Variant & Max gap & Wt.\ gap & $95\%$ CI (Wt.) & Bins & Min/$q_{25}$/Med & Null $q_{.975}$ & Perm.\ $p$ \\
\midrule
Block-AR & $0.3455$ & $0.016$ & $[0.012,\,0.030]$ & $12/15$ & $11/86/143$ & $0.379$ & $\mathbf{0.0382}$ \\
\bottomrule
\end{tabular}
\end{table}

% AUTO-GENERATION GUARD
% This file is hand-curated (multi-seed mean$\pm$std form, 3 seeds, OT2D
% dropped because its column-by-column values are bit-identical to
% Conf+PredEntropy in every seed). The cron-driven make_paper_tables.py
% writes a single-seed alternative form and will clobber this file on
% the next tick if writable. The file should be kept chmod 444 (or
% make_paper_tables.py patched to skip T1) until submission.
\begin{table}[t]
\centering
\setlength{\tabcolsep}{2pt}
\small
\caption{\textbf{Descriptive 14-method benchmark on the Block-AR $b=2$
retrain family} (Swin-Tiny, CIFAR-10; 5,000-example test half; three training
seeds). ECE-style metrics are mean\,$\pm$\,$1$ std; NLL and Brier are seed
means, with paired deltas in Tab.~\ref{tab:paired-delta}. Bold and underline
mark the first two descriptive means; worst-tertile ECE uses
$r_{\mathrm{std}}$.}
\label{tab:calibration_block_ar}
\begin{tabular}{p{2.6cm}cccccc}
\toprule
Method & ECE $\downarrow$ & AdaECE $\downarrow$ & NLL $\downarrow$ & Brier $\downarrow$ & Worst-tert.\ ECE $\downarrow$ & $\Delta$Acc@$1$ (\%) \\
\midrule
No calibration & $0.1514\,{\scriptstyle\pm 0.0028}$ & $0.1514\,{\scriptstyle\pm 0.0028}$ & $0.427$ & $0.175$ & $0.1581\,{\scriptstyle\pm 0.0049}$ & $0.00\,{\scriptstyle\pm 0.00}$ \\
Temp. Scaling (TS) & $0.0111\,{\scriptstyle\pm 0.0008}$ & $0.0123\,{\scriptstyle\pm 0.0004}$ & $0.295$ & $0.146$ & $0.0198\,{\scriptstyle\pm 0.0020}$ & $0.00\,{\scriptstyle\pm 0.00}$ \\
Ensemble TS & $0.0111\,{\scriptstyle\pm 0.0008}$ & $0.0123\,{\scriptstyle\pm 0.0004}$ & $0.295$ & $0.146$ & $0.0198\,{\scriptstyle\pm 0.0020}$ & $0.00\,{\scriptstyle\pm 0.00}$ \\
Vector Scaling & $0.0094\,{\scriptstyle\pm 0.0022}$ & $0.0106\,{\scriptstyle\pm 0.0025}$ & $\mathbf{0.288}$ & $\mathbf{0.142}$ & $0.0169\,{\scriptstyle\pm 0.0022}$ & $+0.24\,{\scriptstyle\pm 0.16}$ \\
Classwise TS & $0.0098\,{\scriptstyle\pm 0.0010}$ & $0.0115\,{\scriptstyle\pm 0.0004}$ & $0.294$ & $0.145$ & $0.0182\,{\scriptstyle\pm 0.0009}$ & $0.00\,{\scriptstyle\pm 0.00}$ \\
Parametric TS & $0.0113\,{\scriptstyle\pm 0.0010}$ & $\underline{0.0094\,{\scriptstyle\pm 0.0005}}$ & $0.298$ & $0.146$ & $0.0179\,{\scriptstyle\pm 0.0007}$ & $0.00\,{\scriptstyle\pm 0.00}$ \\
Histogram Binning & $0.0100\,{\scriptstyle\pm 0.0016}$ & $0.0099\,{\scriptstyle\pm 0.0026}$ & $0.323$ & $0.150$ & $0.0206\,{\scriptstyle\pm 0.0045}$ & $-0.01\,{\scriptstyle\pm 0.01}$ \\
BBQ & $\mathbf{0.0075\,{\scriptstyle\pm 0.0004}}$ & $0.0127\,{\scriptstyle\pm 0.0036}$ & $0.324$ & $0.150$ & $\mathbf{0.0153\,{\scriptstyle\pm 0.0017}}$ & $0.00\,{\scriptstyle\pm 0.00}$ \\
Isotonic Regression & $0.0108\,{\scriptstyle\pm 0.0020}$ & $0.0111\,{\scriptstyle\pm 0.0017}$ & $0.323$ & $0.150$ & $0.0195\,{\scriptstyle\pm 0.0025}$ & $-0.01\,{\scriptstyle\pm 0.01}$ \\
SB-ECE TS & $0.0105\,{\scriptstyle\pm 0.0010}$ & $0.0109\,{\scriptstyle\pm 0.0008}$ & $0.295$ & $0.145$ & $0.0184\,{\scriptstyle\pm 0.0041}$ & $0.00\,{\scriptstyle\pm 0.00}$ \\
\midrule
MEC & $0.0100\,{\scriptstyle\pm 0.0012}$ & $0.0101\,{\scriptstyle\pm 0.0012}$ & $\underline{0.293}$ & $\underline{0.145}$ & $0.0195\,{\scriptstyle\pm 0.0010}$ & $0.00\,{\scriptstyle\pm 0.00}$ \\
Conf-only & $0.0109\,{\scriptstyle\pm 0.0016}$ & $0.0099\,{\scriptstyle\pm 0.0018}$ & $0.299$ & $0.146$ & $0.0180\,{\scriptstyle\pm 0.0007}$ & $0.00\,{\scriptstyle\pm 0.00}$ \\
Conf + PredEntropy & $\underline{0.0085\,{\scriptstyle\pm 0.0006}}$ & $\mathbf{0.0090\,{\scriptstyle\pm 0.0025}}$ & $0.299$ & $0.146$ & $\underline{0.0168\,{\scriptstyle\pm 0.0016}}$ & $0.00\,{\scriptstyle\pm 0.00}$ \\
\textbf{AR-CondCal} & $0.0100\,{\scriptstyle\pm 0.0010}$ & $0.0100\,{\scriptstyle\pm 0.0017}$ & $0.306$ & $0.148$ & $0.0194\,{\scriptstyle\pm 0.0033}$ & $0.00\,{\scriptstyle\pm 0.00}$ \\
\bottomrule
\end{tabular}
\end{table}

\FloatBarrier

\section{Reliability diagrams on Block Attention Residual}
\label{app:reliability}

\begin{figure}[h]
  \centering
  \includegraphics[width=\linewidth]{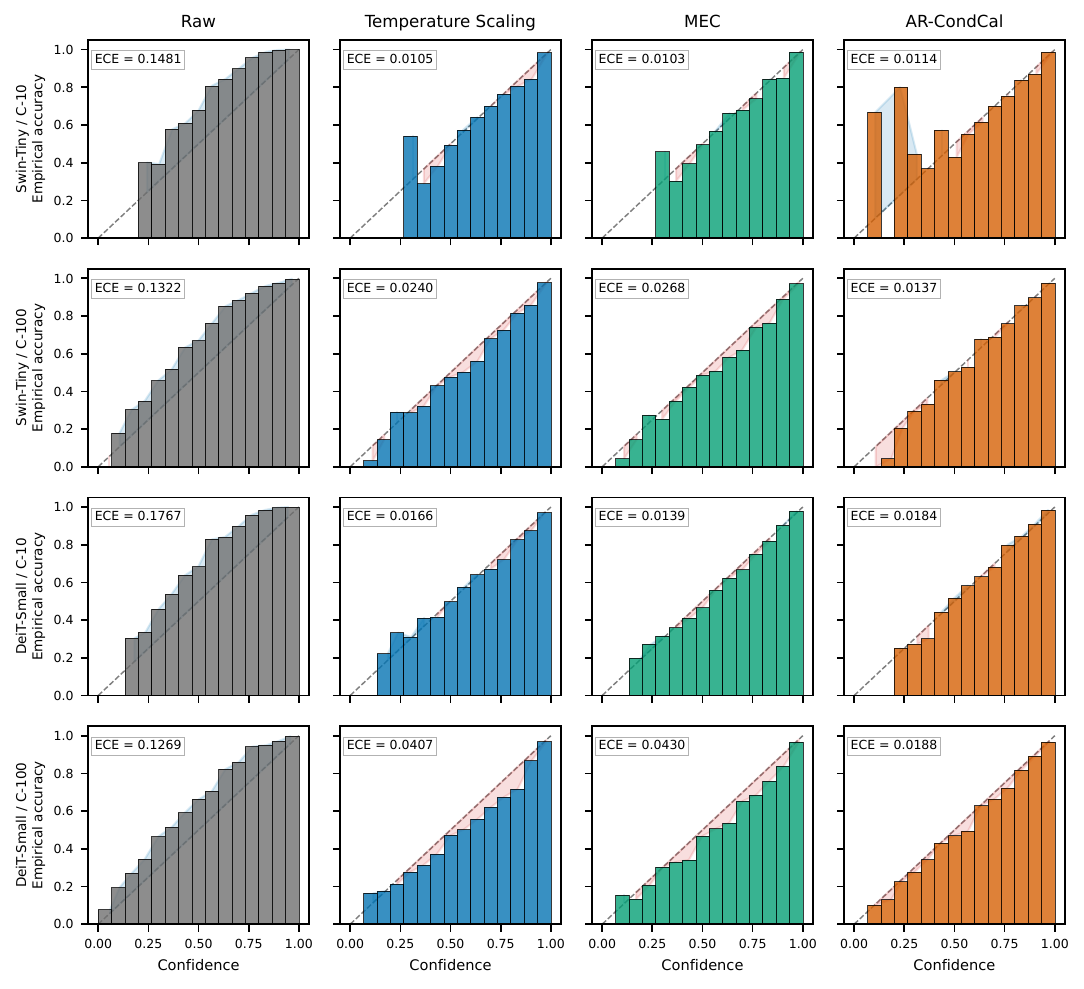}
  \caption{Reliability diagrams on Block-AR for four substrates (rows:
    Swin-Tiny / DeiT-Small $\times$ CIFAR-10 / CIFAR-100) under four
    calibrators (columns). Bars show empirical accuracy in each of $15$
    equal-width confidence bins; dashed line is perfect calibration; red
    shading marks over-confidence (predicted $>$ accuracy) and blue
    marks under-confidence, both relative to the bin's mean confidence.
    Per-panel global ECE is annotated in the upper-left corner. All
    cells use a seed-$0$ ep\,$=$\,$299$ checkpoint with the 50/50 cal/test
    split (seed $42$; the plotted half has $n=5{,}000$). On these uncalibrated
    panels the shading is predominantly blue. All four rows are seed-0 runs of the
    24-run design, with calibrators refitted on their cached logits. The figure is
    descriptive and is not an input to any primary result.}
  \label{fig:reliability-block}
\end{figure}

Figure~\ref{fig:reliability-block} shows reliability diagrams ($15$
equal-width confidence bins, seed $0$, ep\,$=$\,$299$, 50/50 split) for
the four Block-AR cells (rows) and four calibration settings (columns):
uncalibrated, Temperature Scaling, MEC, and AR-CondCal. Each panel
is annotated with its own global ECE; per-bar height is empirical
accuracy at that confidence bin. These checkpoints are systematically
\emph{under}-confident before calibration, not over-confident: on the plotted
test half, accuracy minus mean confidence is $+0.1481$, $+0.1321$, $+0.1767$
and $+0.1269$ for the four rows, and accuracy exceeds mean confidence in
$13/13$, $14/15$, $13/13$ and $15/15$ of the populated bins. Because the
signed bin gaps are almost all of one sign, each panel's ECE coincides with
$|\text{accuracy}-\text{mean confidence}|$ to four decimals. The resolved
training configurations of all four runs have mixup $0.8$, CutMix $1.0$ and
label smoothing $0.1$ active; these are a known route to under-confidence. The
runs also carry a soft-binned ECE auxiliary loss (App.~\ref{app:details}). No
ablation here isolates the cause of the under-confidence. All three calibrators reduce the
resulting gap. The seed-$0$ ordering is
substrate-dependent: AR-CondCal has lower global-ECE point estimates on
the two CIFAR-100 cells but is not uniformly superior across the four
substrates. These single-seed reliability diagrams are descriptive and
motivate the cross-seed controlled analyses in the main text.

\section{Calibration-set-size sensitivity}
\label{app:calsize}

\begin{figure}[h]
  \centering
  \includegraphics[width=0.62\linewidth]{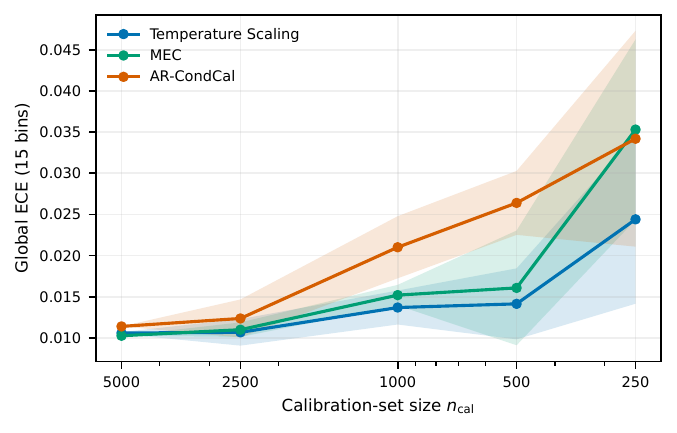}
  \caption{Calibration-set-size sensitivity for the Swin-Tiny/CIFAR-10 Block-AR
    ($b{=}2$) seed-0 run. Curves show mean $\pm$ 1 std across 5 calibration
    subsamples per size with the 5,000-example test split held fixed.}
  \label{fig:calsize}
\end{figure}

We re-fit each calibrator on sub-samples of the 5000-sample calibration
split, holding the 5000-sample test split fixed. Each size is repeated
5 times with different random calibration subsets. Figure
\ref{fig:calsize} reports mean global ECE with $\pm$1-std shaded bands.

Among the three methods examined here, Temperature Scaling
is the most sample-efficient on this substrate: its ECE rises from $0.0106$ at
$n_\text{cal}=5000$ (the mean over five row orders of the same calibration
set; the temperature fit is numerically order-sensitive at the fourth decimal,
which is why a single fit gives $0.0105$ in Fig.~\ref{fig:reliability-block}) to $0.0244$ at $n_\text{cal}=250$, a factor of
$\approx2.3$. This is consistent with the intuition that a single scalar
calibrator parameter can be estimated from relatively few calibration examples
in this setting. Both two-dimensional methods degrade more steeply over the
same range: MEC by a factor of $\approx3.4$ ($0.0103 \to 0.0353$) and
AR-CondCal by $\approx3.0$ ($0.0114 \to 0.0342$). AR-CondCal is therefore not
the more sample-robust of the two-dimensional methods on this run,
and we make no robustness claim in its favour here. At
$n_\text{cal} \le 1000$, their mean curves overlap within the
reported $\pm$1-std variability bands. Between 2,500 and 5,000 examples the mean ECE
changes by $0.00008$ (TS), $0.00074$ (MEC) and $0.00097$ (AR-CondCal). This comparison characterises
sample-size sensitivity among these three methods; the matched test of
routing value is reported in the main text.

\section{MCE, Classwise ECE, and SmoothECE (complement to Table~\ref{tab:calibration_block_ar})}
\label{app:cwece}

Table~\ref{tab:cwece-supp} reports three complementary metrics: MCE (worst-bin
$|\mathrm{acc}-\mathrm{conf}|$ over $15$ equal-width bins), classwise
ECE~\citep{kull2019beyond} (mean of per-class ECE over the $10$
CIFAR-10 classes), and SmoothECE~\citep{blasiok2023smooth}. SmoothECE
and equal-width ECE produce different method orderings, with several
methods moving substantially between them. Proper-score comparisons in
the main text therefore carry the inferential weight. AR-CondCal is not
uniformly best across the alternative ECE summaries.

% Generated by scripts/probe_repair/repair_appendix_e.py; do not hand edit.
\begin{table}[!ht]
\centering
\small
\caption{Additional calibration metrics on the same Block-AR $b=2$
Swin-Tiny/CIFAR-10 family as Table~\ref{tab:calibration_block_ar}.
Every entry is the mean $\pm$ one sample standard deviation over training
seeds 0, 1, and 2, using their fixed 5,000-example evaluation halves.
MCE is the largest accuracy--confidence gap among 15 equal-width bins with
at least five examples. Classwise ECE averages the ten one-versus-rest ECEs.
SmoothECE is the kernel-smoothed metric. Rankings are descriptive and are
not used for inference.}
\label{tab:cwece-supp}
\setlength{\tabcolsep}{4pt}
\begin{tabular}{lccc}
\toprule
Method & MCE $\downarrow$ & Classwise ECE $\downarrow$ & SmoothECE $\downarrow$ \\
\midrule
No calibration & $0.2441\,{\scriptstyle\pm 0.0019}$ & $0.0325\,{\scriptstyle\pm 0.0010}$ & $0.1512\,{\scriptstyle\pm 0.0035}$ \\
Temp. Scaling (TS) & $0.1406\,{\scriptstyle\pm 0.0828}$ & $0.0065\,{\scriptstyle\pm 0.0001}$ & $0.0136\,{\scriptstyle\pm 0.0002}$ \\
Ensemble TS & $0.1406\,{\scriptstyle\pm 0.0828}$ & $0.0065\,{\scriptstyle\pm 0.0001}$ & $0.0136\,{\scriptstyle\pm 0.0002}$ \\
Vector Scaling & $0.0997\,{\scriptstyle\pm 0.0628}$ & $0.0052\,{\scriptstyle\pm 0.0001}$ & $0.0128\,{\scriptstyle\pm 0.0007}$ \\
Classwise TS & $0.1046\,{\scriptstyle\pm 0.0214}$ & $0.0061\,{\scriptstyle\pm 0.0002}$ & $0.0136\,{\scriptstyle\pm 0.0003}$ \\
Parametric TS & $0.0930\,{\scriptstyle\pm 0.0605}$ & $0.0065\,{\scriptstyle\pm 0.0002}$ & $0.0142\,{\scriptstyle\pm 0.0007}$ \\
Histogram Binning & $0.0795\,{\scriptstyle\pm 0.0433}$ & $0.0084\,{\scriptstyle\pm 0.0002}$ & $0.0101\,{\scriptstyle\pm 0.0015}$ \\
BBQ & $0.0366\,{\scriptstyle\pm 0.0130}$ & $0.0080\,{\scriptstyle\pm 0.0002}$ & $0.0081\,{\scriptstyle\pm 0.0006}$ \\
Isotonic Regression & $0.2293\,{\scriptstyle\pm 0.2084}$ & $0.0085\,{\scriptstyle\pm 0.0002}$ & $0.0120\,{\scriptstyle\pm 0.0011}$ \\
SB-ECE TS & $0.0900\,{\scriptstyle\pm 0.0401}$ & $0.0066\,{\scriptstyle\pm 0.0004}$ & $0.0128\,{\scriptstyle\pm 0.0003}$ \\
MEC & $0.1208\,{\scriptstyle\pm 0.0276}$ & $0.0063\,{\scriptstyle\pm 0.0004}$ & $0.0124\,{\scriptstyle\pm 0.0002}$ \\
NW (Conf only) & $0.1661\,{\scriptstyle\pm 0.1028}$ & $0.0067\,{\scriptstyle\pm 0.0001}$ & $0.0115\,{\scriptstyle\pm 0.0013}$ \\
NW (Conf $+$ Pred. Ent.) & $0.1515\,{\scriptstyle\pm 0.0461}$ & $0.0069\,{\scriptstyle\pm 0.0003}$ & $0.0113\,{\scriptstyle\pm 0.0010}$ \\
AR-CondCal & $0.4084\,{\scriptstyle\pm 0.2132}$ & $0.0066\,{\scriptstyle\pm 0.0005}$ & $0.0120\,{\scriptstyle\pm 0.0013}$ \\
\bottomrule
\end{tabular}
\end{table}

% ============================================================
\section{Paired NLL and Brier deltas vs uncalibrated model}
\label{app:paired-delta}

Tab.~\ref{tab:calibration_block_ar} reports seed-mean NLL and Brier;
the paired view here subtracts each seed's uncalibrated score before
aggregation to isolate the within-seed calibrator-induced change. For
the three $b=2$ Block-AR checkpoints, raw NLL is $0.42519$,
$0.43120$, and $0.42348$ (mean $0.42663$, sample std $0.00406$). These
values are distinct from the
two default-block-setting checkpoints described in Appendix~\ref{app:cross-arch-multiseed}.
Tab.~\ref{tab:paired-delta} reports the paired
$\Delta\mathrm{NLL}_{m,s} = \mathrm{NLL}_{m,s} -
\mathrm{NLL}_{\mathrm{raw},s}$ and $\Delta\mathrm{Brier}_{m,s}$,
aggregated as mean\,$\pm$\,$1$ sample std over the $3$ seeds. The
$\Delta$-stds are method-dependent; pairing defines the within-seed
contrast but is not presented as a uniform variance reduction. Vector
Scaling delivers the strongest paired improvement
on both proper scores, followed by MEC and Classwise TS;
AR-CondCal achieves $\Delta\mathrm{NLL}=-0.1202\pm0.0056$, a
$0.0185$ smaller mean improvement than Vector Scaling.

% AUTO-GENERATION GUARD
% Hand-built paired-delta NLL/Brier appendix table. Source:
% c10-swin-ar_block-s{0,1,2}_blocksize2/calibration_benchmark.json.
% Re-audited 2026-08-17.
% Per-seed delta = metric(method, seed) - metric(raw, seed),
% then mean +/- std over the 3 seeds.
\begin{table}[!ht]
\centering
\setlength{\tabcolsep}{6pt}
\small
\caption{\textbf{Paired proper-score changes relative to the
uncalibrated model} (Block-AR, $b=2$, Swin-Tiny, CIFAR-10, ep $=$ $299$;
complement to Tab.~\ref{tab:calibration_block_ar}). For each
training seed $s$, we compute $\Delta\mathrm{NLL}_{m,s} =
\mathrm{NLL}_{m,s} - \mathrm{NLL}_{\mathrm{raw},s}$ and analogously
for Brier, then report mean\,$\pm$\,$1$ std over seeds. This paired
view isolates each calibrator's within-seed change. The raw per-seed
NLLs are $0.42519$, $0.43120$, and $0.42348$ (sample std $0.00406$);
the paired $\Delta$-stds are method-dependent and are not claimed to be
uniformly smaller. \textbf{Bold} = most-negative (best) mean; \underline{underlined}
= $2$nd-best; rank markers use unrounded means.}
\label{tab:paired-delta}
\begin{tabular}{lcc}
\toprule
Method & $\Delta$NLL vs raw $\downarrow$ & $\Delta$Brier vs raw $\downarrow$ \\
\midrule
Temp. Scaling (TS) & $-0.1320\,{\scriptstyle\pm 0.0050}$ & $-0.0298\,{\scriptstyle\pm 0.0016}$ \\
Ensemble TS & $-0.1320\,{\scriptstyle\pm 0.0050}$ & $-0.0299\,{\scriptstyle\pm 0.0016}$ \\
Vector Scaling & $\mathbf{-0.1387\,{\scriptstyle\pm 0.0051}}$ & $\mathbf{-0.0331\,{\scriptstyle\pm 0.0019}}$ \\
Classwise TS & $-0.1328\,{\scriptstyle\pm 0.0049}$ & $-0.0304\,{\scriptstyle\pm 0.0016}$ \\
Parametric TS & $-0.1272\,{\scriptstyle\pm 0.0044}$ & $-0.0284\,{\scriptstyle\pm 0.0011}$ \\
Histogram Binning & $-0.1029\,{\scriptstyle\pm 0.0068}$ & $-0.0259\,{\scriptstyle\pm 0.0014}$ \\
BBQ & $-0.1029\,{\scriptstyle\pm 0.0029}$ & $-0.0252\,{\scriptstyle\pm 0.0009}$ \\
Isotonic Regression & $-0.1034\,{\scriptstyle\pm 0.0067}$ & $-0.0259\,{\scriptstyle\pm 0.0014}$ \\
SB-ECE TS & $-0.1312\,{\scriptstyle\pm 0.0048}$ & $-0.0301\,{\scriptstyle\pm 0.0016}$ \\
MEC & $\underline{-0.1334\,{\scriptstyle\pm 0.0050}}$ & $\underline{-0.0304\,{\scriptstyle\pm 0.0015}}$ \\
Conf-only & $-0.1269\,{\scriptstyle\pm 0.0028}$ & $-0.0293\,{\scriptstyle\pm 0.0018}$ \\
Conf $+$ PredEntropy & $-0.1279\,{\scriptstyle\pm 0.0052}$ & $-0.0293\,{\scriptstyle\pm 0.0019}$ \\
\textbf{AR-CondCal} & $-0.1202\,{\scriptstyle\pm 0.0056}$ & $-0.0278\,{\scriptstyle\pm 0.0025}$ \\
\bottomrule
\end{tabular}
\end{table}

% ============================================================
\section{Robustness checks for the routing-conditional gap}
\label{app:gap-robustness}

This section reports robustness checks for the bin-conditional routing gap on
Block-AR ($0.3455$) reported in
Tab.~\ref{tab:phenomenon} (Swin-Tiny + CIFAR-10 seed-$0$). All
checks are computed on the same full $10{,}000$-sample CIFAR-10
test population, with the same $15$ equal-width confidence bins,
routing entropy, and the same ``$n \ge 5$ samples per shared bin''
rule used by the diagnostic pipeline. They characterise sampling variation,
bin support, and sensitivity to the discrete conditioning scheme.

\textbf{Bootstrap CI for the max-gap statistic.}
Resampling the $10{,}000$ test instances with replacement
($B = 5000$ resamples, fixed tertile cuts $\mathrm{q33}, \mathrm{q67}$
re-applied per replicate) gives the $95\%$ percentile bootstrap CIs
in Tab.~\ref{tab:gap-robustness}: Block-AR max-gap CI
$[0.109, 0.678]$. This interval is
wide, reflecting the known instability of the max-over-bins
statistic under finite-sample resampling; the Block-AR
point estimate ($0.3455$) sits in the lower half of its CI,
consistent with the high variance of the
max-over-bins statistic under a per-bin minimum-sample-count
rule ($n \ge 5$). The CIs are reported as a robustness summary
of the sampling distribution; we use the permutation test below
for the binned significance assessment.

\textbf{Within-confidence-bin permutation test.}
We test whether routing-tertile membership carries information about correctness
once confidence-bin membership is fixed. Inside each of the 15 confidence bins we
permute $r_{\rm agg}$ across examples ($B=5{,}000$ permutations), recompute the
max-gap statistic and report $p=(1+\#\{\text{null}\ge\text{observed}\})/(B+1)$. The
per-run pipeline draws its bootstrap resamples and then its permutations from one
random stream seeded with 42. Re-executing it reproduces every stored value exactly;
for this run $p=0.0382$ (190 exceedances), and its sibling seeds give $0.4771$ and
$0.6989$. We treat the rejection as an unreplicated observation; it does not
establish a routing-defined subgroup effect across seeds. Proposition~2 of
App.~\ref{app:theory} motivates looking for such an effect at fixed confidence; the
binned statistic only approximates that comparison and is sensitive to finite
support.

\textbf{Borderline run.} DeiT-Small Full-AR on CIFAR-100 seed 1 gives $p=0.0494$
(246 exceedances of 5,000; Monte Carlo standard error $0.0031$), reproduced
exactly by re-execution. Five further recomputations with 50,000 permutations each
and fixed seeds 1--5 give $0.0498$, $0.0484$, $0.0503$, $0.0502$ and $0.0479$ (mean
$0.0493$). The exact permutation $p$-value is therefore within Monte Carlo error of
$0.05$, and we call the run borderline rather than a second clear rejection. Its
sibling seeds give $0.9268$ and $0.7471$.

\textbf{Binned-test validity.}
To assess sensitivity to coarse bins, we preserve the observed $(c,r)$ pairs, draw
correctness under cross-fitted confidence-only nulls (isotonic and logistic
spline), and repeat the binned procedure with 10, 15, 20, 30 and 50 bins. Across
four settings, 2,000 simulations per setting and 199 permutations per simulation,
the largest Type-I estimate is $0.0545$ at nominal $0.05$ (Monte Carlo standard
error about $0.005$). The fitted nulls reveal no detectable anti-conservative
behaviour, although the binned test remains distinct from exact conditioning on
continuous $c$.

\textbf{Continuous-confidence check.} The check was first run on four
representative runs (Swin-Tiny CIFAR-10 Block and Full, DeiT-Small CIFAR-10 Block,
Swin-Tiny CIFAR-100 Block, all seed 0). After the borderline result we extended it,
with the same estimators, folds and seeds, to all 24 replicated runs; the extension
reproduces the four stored results to within $10^{-15}$
(Table~\ref{tab:continuous-24}). Five-fold cross-fitted logistic correctness models
on $\operatorname{logit}(c)$ are compared with the same models plus standardised
$r_{\rm agg}$ or plus the whole standardised profile. Adding $r_{\rm agg}$ never
reduces held-out log-loss with an interval excluding zero; three runs get worse,
and all changes are below $0.5\times10^{-3}$ nats. Adding the profile reduces
log-loss in one run (Swin-Tiny Full-AR, CIFAR-100, seed 1, by $1.7\times10^{-3}$
nats) and increases it in one; the improved run's siblings do not repeat the
reduction. The three DeiT-Small Full-AR CIFAR-100 seeds, including the borderline
run, have intervals that include zero for both augmentations. These 48 intervals
are per-run example bootstraps without multiplicity correction; the one improvement
is uncorrected and not repeated across seeds, and neither it nor the number of
intervals excluding zero is evidence about the joint null.

Local-caliper tests permute $r_{\rm agg}$ within consecutive confidence-sorted
windows of 50, 100, 200 and 400 examples and measure the mean absolute within-window
correlation between $r_{\rm agg}$ and correctness. Across the 96 run--window
combinations, 5 give $p<0.05$ (4.8 expected under the null): windows 100, 200 and 400
of DeiT-Small Block-AR CIFAR-100 seed 0 ($p=0.001$, $0.015$, $0.004$), whose sibling
seeds give $p\ge0.168$; window 100 of the borderline run ($p=0.025$); and window 400
of Swin-Tiny Block-AR CIFAR-100 seed 2 ($p=0.037$). With 1,000 permutations the
smallest attainable $p$ is $0.001$, above a Bonferroni threshold of $0.05/96$, so
these tests are descriptive rather than a corrected family, and the absence of
corrected rejections is not evidence of absence.

% Generated by scripts/revision_2026_09_25/continuous_table.py.
\begin{table}[!htbp]
\centering\scriptsize
\caption{Continuous-confidence check on all 24 replicated runs (post-hoc extension of the original four
representative runs). $\Delta L$ is augmented minus confidence-only held-out binary log-loss ($10^{-3}$ nats;
negative favours routing) for five-fold cross-fitted logistic models on $\operatorname{logit}(c)$ plus
$r_{\rm agg}$ or plus the whole profile, with 95\% paired example-bootstrap intervals (2,000 draws); bold
intervals exclude zero. Caliper columns give permutation $p$-values (1,000 permutations, smallest attainable
$p=0.001$) for the association between $r_{\rm agg}$ and correctness within consecutive confidence-sorted windows
of 50, 100, 200 and 400 examples.}
\label{tab:continuous-24}
\setlength{\tabcolsep}{3pt}
\begin{tabular}{lllcccccc}
\toprule
Setting & Variant & Seed & $\Delta L$, $r_{\rm agg}$ & $\Delta L$, profile & $p_{50}$ & $p_{100}$ & $p_{200}$ & $p_{400}$ \\
\midrule
Swin/C-10 & Block & 0 & $+0.11\;[-0.01,+0.24]$ & $+0.34\;[-0.50,+1.22]$ & 0.500 & 0.625 & 0.653 & 0.798 \\
Swin/C-10 & Block & 1 & $+0.00\;[-0.10,+0.10]$ & $-0.63\;[-1.60,+0.45]$ & 0.682 & 0.254 & 0.421 & 0.913 \\
Swin/C-10 & Block & 2 & $+0.01\;[-0.11,+0.13]$ & $+0.16\;[-0.54,+0.86]$ & 0.774 & 0.623 & 0.491 & 0.083 \\
Swin/C-10 & Full & 0 & $-0.12\;[-0.53,+0.29]$ & $-0.02\;[-1.27,+1.28]$ & 0.135 & 0.331 & 0.206 & 0.063 \\
Swin/C-10 & Full & 1 & $-0.30\;[-0.77,+0.17]$ & $+0.42\;[-0.94,+1.75]$ & 0.434 & 0.740 & 0.329 & 0.345 \\
Swin/C-10 & Full & 2 & $-0.14\;[-0.48,+0.19]$ & $-1.41\;[-2.81,+0.09]$ & 0.831 & 0.523 & 0.296 & 0.679 \\
\addlinespace
Swin/C-100 & Block & 0 & $-0.41\;[-1.02,+0.15]$ & $-0.93\;[-2.16,+0.28]$ & 0.519 & 0.571 & 0.520 & 0.178 \\
Swin/C-100 & Block & 1 & \textbf{$+0.12\;[+0.00,+0.23]$} & $+0.32\;[-0.73,+1.37]$ & 0.857 & 0.951 & 0.570 & 0.224 \\
Swin/C-100 & Block & 2 & \textbf{$+0.17\;[+0.05,+0.31]$} & $-0.22\;[-1.25,+0.82]$ & 0.076 & 0.364 & 0.469 & 0.037 \\
Swin/C-100 & Full & 0 & $-0.39\;[-0.98,+0.21]$ & $-0.69\;[-1.96,+0.65]$ & 0.280 & 0.520 & 0.388 & 0.439 \\
Swin/C-100 & Full & 1 & $-0.03\;[-0.23,+0.17]$ & \textbf{$-1.72\;[-3.12,-0.29]$} & 0.963 & 0.997 & 0.983 & 0.966 \\
Swin/C-100 & Full & 2 & $+0.03\;[-0.16,+0.22]$ & $+0.68\;[-0.37,+1.72]$ & 0.213 & 0.167 & 0.608 & 0.436 \\
\addlinespace
DeiT/C-10 & Block & 0 & \textbf{$+0.23\;[+0.06,+0.39]$} & \textbf{$+1.15\;[+0.01,+2.36]$} & 0.127 & 0.173 & 0.101 & 0.114 \\
DeiT/C-10 & Block & 1 & $+0.03\;[-0.35,+0.41]$ & $-0.59\;[-2.35,+1.17]$ & 0.720 & 0.685 & 0.345 & 0.380 \\
DeiT/C-10 & Block & 2 & $+0.15\;[-0.23,+0.56]$ & $-0.88\;[-2.55,+0.82]$ & 0.198 & 0.425 & 0.583 & 0.558 \\
DeiT/C-10 & Full & 0 & $-0.01\;[-0.31,+0.30]$ & $+1.09\;[-0.37,+2.45]$ & 0.756 & 0.257 & 0.345 & 0.749 \\
DeiT/C-10 & Full & 1 & $+0.08\;[-0.26,+0.41]$ & $+0.04\;[-1.73,+1.73]$ & 0.143 & 0.181 & 0.070 & 0.060 \\
DeiT/C-10 & Full & 2 & $-0.40\;[-1.01,+0.23]$ & $-0.10\;[-1.77,+1.48]$ & 0.243 & 0.144 & 0.097 & 0.120 \\
\addlinespace
DeiT/C-100 & Block & 0 & $-0.10\;[-0.60,+0.39]$ & $-1.27\;[-2.96,+0.35]$ & 0.339 & 0.001 & 0.015 & 0.004 \\
DeiT/C-100 & Block & 1 & $-0.05\;[-0.35,+0.21]$ & $-0.39\;[-1.97,+1.23]$ & 0.227 & 0.493 & 0.313 & 0.168 \\
DeiT/C-100 & Block & 2 & $-0.15\;[-0.61,+0.32]$ & $-0.99\;[-2.79,+0.81]$ & 0.496 & 0.432 & 0.175 & 0.184 \\
DeiT/C-100 & Full & 0 & $-0.27\;[-0.82,+0.29]$ & $-0.64\;[-2.34,+1.02]$ & 0.150 & 0.265 & 0.213 & 0.186 \\
DeiT/C-100 & Full & 1 & $-0.39\;[-1.08,+0.32]$ & $-0.61\;[-2.52,+1.18]$ & 0.263 & 0.025 & 0.665 & 0.202 \\
DeiT/C-100 & Full & 2 & $-0.06\;[-0.40,+0.29]$ & $+0.99\;[-0.34,+2.25]$ & 0.295 & 0.849 & 0.586 & 0.499 \\
\bottomrule
\end{tabular}
\end{table}

\textbf{Bin-support diagnostics.}
The max-gap and the weighted-integrated gap below are computed
over the bins shared by the low- and high-routing tertiles.
Tab.~\ref{tab:gap-robustness} reports the number of shared bins
and the per-shared-bin sample-count statistics
($\min$, $25^{\text{th}}$, median); Block-AR shares $12$ of $15$
bins. Their reported sample counts meet the minimum-support
rule, although satisfying that rule does not eliminate sampling variability.

\textbf{Weighted integrated gap.}
As a summary across bins, we report a weighted mean
$|a_{\mathrm{low}}(c) - a_{\mathrm{high}}(c)|$ across shared bins,
weighted by $w_c = \min(n_{\mathrm{low}}^c, n_{\mathrm{high}}^c)$.
The Block-AR point estimate is $0.016$ with 95\%
percentile-bootstrap CI $[0.012,0.030]$. The much smaller absolute
scale relative to the max-gap shows that most shared bins have far
smaller binned differences than the extreme bin. The weighted gap is
descriptive; the sweep-level inference uses the max-gap permutation
tests in Tab.~\ref{tab:ar-sweep}.

\begin{table}[!ht]
\centering
\setlength{\tabcolsep}{4pt}
\small
\caption{Routing-conditional max-gap robustness for Swin-Tiny /
CIFAR-10 Block-AR seed 0. Intervals use 5000 bootstrap resamples;
the permutation $p$-value uses 5000 within-bin permutations.}
\label{tab:gap-robustness}
\begin{tabular}{lr}
\toprule
Statistic & Block-AR \\
\midrule
Max gap (point) & $0.3455$ \\
\hspace{1em} $95\%$ bootstrap CI & $[0.109,\,0.678]$ \\
Weighted integrated gap (point) & $0.016$ \\
\hspace{1em} $95\%$ bootstrap CI & $[0.012,\,0.030]$ \\
Within-bin permutation $p$ & $\mathbf{0.0382}$ \\
Shared bins (of $15$) & $12$ \\
\hspace{1em} min/p25/median samples per shared bin & $11$/$86$/$143$ \\
\bottomrule
\end{tabular}
\end{table}

\begin{table}[!ht]
\centering
\setlength{\tabcolsep}{4pt}
\footnotesize
\setlength{\tabcolsep}{2pt}
\caption{Per-seed Swin-Tiny / CIFAR-10 diagnostic statistics for the
three-run Block-AR comparison in Fig.~\ref{fig:cross-run-stability}.
All rows use the same 10,000-example test set and diagnostic protocol;
Wt.\ CI is a percentile-bootstrap interval.}
\label{tab:perseed-swin-c10-diagnostic}
\begin{tabular}{llcccccccc}
\toprule
Variant & Seed & Acc@1 & Max gap & Wt.\ gap & Wt.\ CI & Bins & Min/$q_{25}$/Med & Perm.\ $p$ & Reject? \\
\midrule
Block-AR ($b\!=\!2$) & $0$ & $0.903$ & $0.3455$ & $0.016$ & $[0.012,\,0.030]$ & $12/15$ & $11/86/143$  & $\mathbf{0.0382}$ & yes \\
Block-AR ($b\!=\!2$) & $1$ & $0.901$ & $0.1325$ & $0.011$ & $[0.011,\,0.028]$ & $12/15$ & $18/117/165$ & $0.4771$ & no  \\
Block-AR ($b\!=\!2$) & $2$ & $0.902$ & $0.1000$ & $0.017$ & $[0.013,\,0.032]$ & $12/15$ & $25/134/171$ & $0.6989$ & no  \\
\bottomrule
\end{tabular}
\end{table}

Every statistic in Tables~\ref{tab:phenomenon}, \ref{tab:gap-robustness} and
\ref{tab:perseed-swin-c10-diagnostic} comes from the per-run diagnostic output, whose
bootstrap intervals and permutation $p$-value share one seed-42 random stream and are
reproduced exactly by re-execution (supplementary material).

\section{Bandwidth sensitivity for AR-CondCal}
\label{app:bw-sensitivity}

\S\ref{sec:rq3} reports AR-CondCal under the default Scott
bandwidth. To check whether the worst-tertile instability is
explained by that fixed bandwidth choice, we sweep eight
seed-level AR runs across five CIFAR-10 substrate / variant
groups: Swin-Tiny~$+$~Block-AR ($3$ seeds),
DeiT-Small~$+$~Block-AR ($1$ seed),
ViT-B/$16$~$+$~Block-AR ($1$ seed),
DeiT-Small~$+$~Full-AR ($2$ seeds), and
ViT-B/$16$~$+$~Full-AR ($1$ seed). These are the CIFAR-10 runs on which the
cross-architecture AR-CondCal benchmark had been completed when the analysis was
run; the selection rule was not recorded in advance, so the results describe these
runs only. We use the same calibration-benchmark population as
Tab.~\ref{tab:calibration_block_ar}: a fixed seed-$42$ $5{,}000/5{,}000$
calibration/test split. This analysis computes worst-tertile ECE with $15$
equal-width bins inside each $r_{\mathrm{std}}$ tertile, whereas the main scalar
analysis uses $10$ (App.~\ref{app:wt-ece-illustration}); its values are therefore not
directly comparable with Fig.~\ref{fig:scalar-effect} or
Table~\ref{tab:calibration_block_ar}.
We compare five bandwidth modes: $0.5\times$, $1\times$,
$2\times$ the per-axis Scott bandwidth; the multiplier from
$\{0.25, 0.5, 1, 2, 4\}$ minimising mean $5$-fold calibration
NLL (\textbf{CV-NLL}); and the multiplier minimising held-out
global ECE (\textbf{Oracle-ECE}, a deliberately optimistic
ceiling, not a selection method).

\begin{table}[h]
\centering\scriptsize
\setlength{\tabcolsep}{1.5pt}
\caption{\textbf{Bandwidth sensitivity for AR-CondCal.} Pooled
mean$\,\pm\,$std across the available eight CIFAR-10 AR
checkpoints described above.}
\label{tab:bw-sensitivity}
\begin{tabular}{lcccc}
\toprule
Mode & Multiplier / selection & Global ECE & Worst-tertile ECE & NLL \\
\midrule
$0.5\!\times\!$ Scott & $0.50$                  & $0.0182\!\pm\!0.0039$ & $0.0288\!\pm\!0.0080$ & $0.4589\!\pm\!0.2307$ \\
$1\!\times\!$ Scott   & $1.00$ (paper default)  & $0.0105\!\pm\!0.0041$ & $0.0235\!\pm\!0.0074$ & $0.4348\!\pm\!0.2343$ \\
$2\!\times\!$ Scott   & $2.00$                  & $0.0159\!\pm\!0.0085$ & $0.0261\!\pm\!0.0102$ & $0.4336\!\pm\!0.2398$ \\
CV-NLL                & $5$-fold cal.\ NLL      & $0.0129\!\pm\!0.0034$ & $0.0244\!\pm\!0.0067$ & $0.4317\!\pm\!0.2361$ \\
Oracle-ECE            & held-out global ECE     & $0.0103\!\pm\!0.0037$ & $0.0232\!\pm\!0.0069$ & $0.4362\!\pm\!0.2372$ \\
\bottomrule
\end{tabular}
\end{table}

\textbf{Bandwidth comparison.} CV-NLL changes worst-tertile ECE by
$+0.0009$ relative to $1\!\times\!$ Scott, well within the
cross-run standard deviation of approximately $0.007$; the
global-ECE oracle reduces worst-tertile ECE by $0.0003$ relative
to $1\!\times\!$ Scott, also within the cross-run standard
deviation. No single multiplier is consistently best across the
eight runs. These comparisons provide no support for attributing the
worst-tertile pattern to the default Scott bandwidth alone.

\textbf{Caveat.} The audit does not address anisotropic
bandwidth grids, non-Gaussian kernels, or objectives that tune
worst-tertile ECE directly; its conclusion concerns the bandwidth
choices and runs evaluated here.

\section{Pooled effect for the scalar probe}
\label{app:effect-size}
For each of 24 seed-level runs nested in four backbone--dataset settings we compute
$\Delta=\mathrm{WTECE}_{\text{AR-CondCal}}-
\mathrm{WTECE}_{\text{Conf-only}}$. Margins of $0.0025$, $0.005$ and $0.010$ were
specified for this pooled reanalysis, not for the original study; $0.005$ was the
primary margin and $0.0025$ the tight margin. A
setting-clustered hierarchical bootstrap gives pooled mean $+0.000380$,
median $+0.000502$, standard deviation $0.00516$, and 12/0/12
wins/ties/losses. The 90\% interval is
$[-0.001302,+0.002089]$ and the 95\% interval is
$[-0.001677,+0.002449]$. The former is compatible with the tight
$\pm0.0025$ margin under this bootstrap; because only four settings are
available, we do not treat this as a formal equivalence test. The 95\%
interval does not include pooled average improvements larger than about
$1.7\times10^{-3}$; under four settings this is an exploratory,
procedure-specific interval, not population coverage. Individual-run differences
range from $-0.00928$ to $+0.00991$, so the pooled interval must not be read
as a bound for every setting.

The same analysis finds ECE difference $-0.000577$ (95\% CI
$[-0.001788,+0.000586]$) and AdaECE difference $-0.000863$
($[-0.002362,+0.000379]$). Proper-score guardrails favour the
confidence-only control: NLL $+0.008423$
($[+0.006055,+0.010903]$) and Brier $+0.001031$
($[+0.000760,+0.001341]$). These are pooled clustered intervals, not
claims of homogeneous effects across settings.

\textbf{Post-hoc sensitivity to runs with a deviating training history.} Omitting
Swin-Tiny Full-AR and DeiT-Small Block-AR on CIFAR-10 seed 0 (accuracies 0.744
and 0.695), a check specified during this revision after the 24-run results were
known, leaves 22 runs in four settings. The same bootstrap gives worst-tertile ECE
$+0.000585$ $[-0.001607,+0.002709]$, NLL $+0.008916$ $[+0.006198,+0.011524]$ and
Brier $+0.001077$ $[+0.000766,+0.001450]$. Two further subsets were specified after
the training-log audit of App.~\ref{app:details}: omitting the two runs trained with
an effective batch of 64 (DeiT-Small CIFAR-10 seed 0, Block and Full; 22 runs) gives
$+0.000787$ $[-0.001314,+0.002931]$, $+0.008836$ $[+0.006201,+0.011798]$ and
$+0.001056$ $[+0.000758,+0.001467]$, and omitting all three runs whose training
deviated from the standard recipe (21 runs) gives $+0.000561$ $[-0.001612,+0.002864]$,
$+0.008939$ $[+0.006166,+0.011865]$ and $+0.001081$ $[+0.000769,+0.001506]$, for
worst-tertile ECE, NLL and Brier respectively (record-weighted; equal-setting values
are in the supplementary material). The 24-run analysis remains primary.

\section{Entropy-profile correctness prediction}
\label{app:probe-sanity}

The profile analysis of \S\ref{sec:probe} predicts $T=\mathbf{1}[\hat y=y]$ with
five-fold cross-fitting, repeated twice (outer seeds 100 and 101), on the 10,000
test examples of each run. The profile, confidence-only and shuffled arms share
outer folds, inner 80/20 training/validation splits, MLP initialisation seeds and
the same two-candidate budget: the arm's one-hidden-layer MLP with 32 ReLU units, or
a shared regularised logistic model of $\operatorname{logit}(c)$ with $C=1$. The
confidence arm's MLP receives zeros in place of the profile, which matches input
dimension but not effective capacity. Adam uses learning rate $10^{-3}$, weight
decay $10^{-3}$ and batches of 200. Binary validation log-loss selects the epoch
count (at most 300, patience 20, minimum improvement $10^{-5}$) and the candidate;
a fresh MLP is then refitted for the selected epoch count on the outer training
data. Scaling is fitted on inner training data during selection and on outer
training data during refitting. Whole profiles are shuffled independently within
inner training, validation, outer training and outer test partitions. No outer
evaluation label enters preprocessing or selection. Scores average each example's
two out-of-fold probabilities; AUROC is secondary.

Table~\ref{tab:repaired-primary} gives the four setting means and the pooled
estimates. Each of 5,000 bootstrap draws (seed 0) samples four setting labels with
replacement, resamples that setting's six paired runs with replacement for every
selected occurrence, and pools the sampled run differences; percentiles 2.5 and
97.5 form the interval. The draws preserve pairing across arms. With six runs per
setting, record-weighted and equal-setting pooling give identical estimates and
intervals. Four observed clusters do not establish general-population coverage, and
the procedure does not estimate an oracle information bound.

Profile minus confidence-only log-loss stays positive when any one setting is
omitted, with leave-one-setting-out means from $+0.000767$ to $+0.001578$; profile
minus shuffled profiles stays negative in both log-loss and Brier score. Across
720 MLP searches there are no fitting failures or optimiser warnings; three
searches reach the epoch budget, and no selected epoch lies at the cap. The
supplementary material lists all 24 run differences for both proper scores and
AUROC.

\textbf{Post-hoc sensitivity without the two lower-accuracy runs.} Omitting
Swin-Tiny Full-AR and DeiT-Small Block-AR on CIFAR-10 seed 0 leaves 22 runs with
five runs in the Swin/CIFAR-10 and DeiT/CIFAR-10 settings, so record-weighted and
equal-setting pooling now differ; both use the same bootstrap. Profile minus
confidence-only is $+1.41$ $[+0.47,+2.65]$ (record) and $+1.50$ $[+0.47,+2.69]$
(equal-setting) in log-loss, and $+0.34$ $[+0.14,+0.59]$ and $+0.36$ $[+0.15,+0.60]$
in Brier. Profile minus shuffled profiles is $-0.29$ $[-0.54,-0.09]$ and $-0.30$
$[-0.55,-0.09]$ in log-loss; in Brier it is $-0.053$ and $-0.055$, with intervals
whose upper limits ($-0.0004\times10^{-3}$ and $+0.0005\times10^{-3}$) sit at zero.
The orderings are unchanged; the Brier advantage over shuffled profiles is not
resolved in this subset. Omitting instead the two runs trained with an effective
batch of 64 (22 runs), profile minus confidence-only log-loss is $+1.23$
$[+0.46,+2.53]$ (record) and $+1.41$ $[+0.46,+2.73]$ (equal-setting), and profile
minus shuffled profiles $-0.30$ $[-0.57,-0.10]$ and $-0.32$ $[-0.57,-0.11]$; both
Brier intervals against shuffled profiles exclude zero. Omitting all three runs
whose training deviated from the standard recipe (21 runs) gives $+1.34$
$[+0.49,+2.64]$ and $+1.52$ $[+0.50,+2.80]$ against confidence-only and $-0.26$
$[-0.51,-0.08]$ and $-0.28$ $[-0.52,-0.08]$ against shuffled profiles in log-loss;
in Brier the interval against shuffled profiles crosses zero ($-0.051$
$[-0.117,+0.003]$ and $-0.056$ $[-0.120,+0.003]$). None of these subsets isolates an
effect of batch size, learning rate or the auxiliary loss, and the 24-run analysis
remains primary.

% Generated by scripts/revision_2026_09_25/render_revision.py from corrected_24run_2026-09-25/profile.
\begin{table}[!htbp]
\centering\small
\caption{Entropy-profile probe: profile minus comparator in binary log-loss ($10^{-3}$ nats) and
Brier score ($10^{-3}$); negative values favour the profile. Setting rows are means over six runs;
the pooled row is the mean over 24 runs with a 95\% hierarchical-bootstrap interval (settings, then
runs; 5,000 draws). Record-weighted and equal-setting pooling coincide in this balanced design.
Four settings support only procedure-specific intervals.}
\label{tab:repaired-primary}
\setlength{\tabcolsep}{4pt}
\begin{tabular}{lrrrr}
\toprule
& \multicolumn{2}{c}{Log-loss} & \multicolumn{2}{c}{Brier} \\
\cmidrule(lr){2-3}\cmidrule(lr){4-5}
Setting & $-$\,Confidence & $-$\,Shuffled & $-$\,Confidence & $-$\,Shuffled \\
\midrule
Swin / CIFAR-10 & $+1.243$ & $-0.446$ & $+0.296$ & $-0.092$ \\
Swin / CIFAR-100 & $+0.664$ & $-0.164$ & $+0.219$ & $-0.013$ \\
DeiT / CIFAR-10 & $+2.826$ & $-0.425$ & $+0.636$ & $-0.073$ \\
DeiT / CIFAR-100 & $+0.394$ & $-0.160$ & $+0.114$ & $-0.064$ \\
\midrule
Pooled (24 runs) & $+1.282$ & $-0.299$ & $+0.317$ & $-0.061$ \\
95\% interval & $[+0.455,+2.364]$ & $[-0.543,-0.102]$ & $[+0.133,+0.543]$ & $[-0.128,-0.005]$ \\
\bottomrule
\end{tabular}
\end{table}

\subsection{An earlier implementation of the profile probe}
\label{app:profile-repair}
An earlier implementation of this probe used accuracy-based early stopping with
weight restoration, fitted feature scaling on all examples, and drew shuffle donors
across outer-fold boundaries. On the 23 runs analysed at the time, it reported a
profile-minus-confidence log-loss difference of $-0.0199$ (95\% interval
$[-0.0429,+0.0093]$), favouring routing. The procedure above removes these choices;
on the same 23 runs it gives $+0.0013$, and on all 24 runs $+0.0013$
(Table~\ref{tab:repaired-primary}). The earlier tables and figure, the mapping from
earlier to current estimates, and a recomputation of the earlier procedure on all
24 runs are retained in the supplementary audit record; they play no role in the
conclusions.

\subsection{Sensitivity audit of the scalar and profile probes}
\label{app:profile-sensitivity}
The audit uses the Block-AR seed-0 run of each backbone--dataset setting.
Under the explicit local-$Q$ law, a confidence anchor $c_i$ is drawn uniformly
from the cache and a whole profile $R_j$ uniformly from its adjacent-confidence
block (target 100 rows; confidence ties are not split). This preserves the
empirical confidence and profile marginals while locally smoothing their
joint distribution. A fixed logistic-spline $q_0(c)$ defines baseline
correctness. The scalar teachers use within-block ranks of $r_{\rm agg}$ (for the
diagnostic) or of $r_{\rm std}$ (for the NW probe). The profile teacher, fixed before
the audit, is the product of two bounded principal-component scores after
covariate-only residualisation of the profile on confidence, $r_{\rm agg}$ and
$r_{\rm std}$; its transforms are fitted on a fixed development half and were not
selected by observed recovery.
For $T^*\sim\operatorname{Bernoulli}(\eta_\beta(c_i,R_j))$, set
$\eta_\beta(c_i,R_j)=\sigma(a_i(\beta)+\beta g(c_i,R_j))$ and solve each
$a_i$ so that $\mathbb{E}_Q[T^*\mid c]=q_0(c)$. The oracle binary-log-loss gain is
\begin{equation}
G_{\rm LL}=\mathbb{E}_Q\!\left[
D_{\rm KL}\!\left(\operatorname{Bern}(\eta_\beta(c,R))\,
\Vert\,\operatorname{Bern}(q_0(c))\right)\right]
=I_Q(T^*;R\mid c).
\label{eq:pilot-oracle}
\end{equation}
The generator is calibrated to this quantity, not to multiclass NLL.
Expected-label evaluation integrates each held-out binary score over the
known Bernoulli label probability. Realised-label scores are also retained.

Table~\ref{tab:scalar-sensitivity} reports both scalar audits at every strength.
Under the null the binned diagnostic rejects in 3.1--4.8\% of 1,000 simulated
datasets per setting at $\alpha=0.05$ and in 0.1\% at $0.05/30$, so its pre-specified
inflation screen passes. Its power at $0.05/30$ rises from near zero at 0.0005 nats
to 0.185--0.85 at 0.010 nats, depending on the setting. The NW probe is worse than
NW$(c)$ under the null and at 0.0005 and 0.002 nats, and recovers 54--65\% of a
0.010-nat effect. These rates and recoveries do not validate profile-prediction
sensitivity, which uses a different teacher.

% Generated by scripts/revision_2026_09_25/scalar_sensitivity_latex.py.
\begin{table}[!htbp]
\centering\scriptsize
\caption{Scalar sensitivity audit on the Block-AR seed-0 run of each setting, at every injected strength
(oracle binary log-loss gain, nats per example). \emph{Diagnostic}: rejection rate of the binned $r_{\rm agg}$
test (9,999 permutations per simulated dataset) under a teacher built from within-block ranks of $r_{\rm agg}$,
at $\alpha=0.05$ and at the family threshold $0.05/30$, recomputed from per-replicate $p$-values; $n$
simulated datasets. \emph{NW probe}: expected-label gain of NW$(c,r_{\rm std})$ over NW$(c)$ ($10^{-3}$
nats; positive favours routing) under a teacher built from ranks of $r_{\rm std}$, with 95\% Monte Carlo
intervals over 50 replicates, and recovery $=$ gain$/$oracle gain. Rejection rates and recovery are
different quantities and the two teachers differ.}
\label{tab:scalar-sensitivity}
\setlength{\tabcolsep}{3.5pt}
\begin{tabular}{llrrrrr}
\toprule
& Oracle & \multicolumn{3}{c}{Diagnostic ($r_{\rm agg}$)} & \multicolumn{2}{c}{NW probe ($r_{\rm std}$)} \\
\cmidrule(lr){3-5}\cmidrule(lr){6-7}
Setting & gain & $n$ & $p<0.05$ & $p<0.05/30$ & Gain & Recovery \\
\midrule
Swin / CIFAR-10 & 0 & 1000 & 0.042 & 0.001 & $-5.24\;[-5.78,-4.70]$ & -- \\
Swin / CIFAR-10 & 0.0005 & 200 & 0.095 & 0.000 & $-4.41\;[-4.83,-3.99]$ & $-883$\% \\
Swin / CIFAR-10 & 0.002 & 200 & 0.170 & 0.010 & $-2.88\;[-3.43,-2.34]$ & $-144$\% \\
Swin / CIFAR-10 & 0.005 & 200 & 0.545 & 0.035 & $+0.14\;[-0.30,+0.58]$ & $3$\% \\
Swin / CIFAR-10 & 0.010 & 200 & 0.930 & 0.185 & $+5.37\;[+5.03,+5.71]$ & $54$\% \\
\addlinespace
Swin / CIFAR-100 & 0 & 1000 & 0.048 & 0.001 & $-3.36\;[-3.63,-3.10]$ & -- \\
Swin / CIFAR-100 & 0.0005 & 200 & 0.145 & 0.010 & $-2.88\;[-3.11,-2.66]$ & $-577$\% \\
Swin / CIFAR-100 & 0.002 & 200 & 0.645 & 0.160 & $-1.35\;[-1.56,-1.14]$ & $-67$\% \\
Swin / CIFAR-100 & 0.005 & 200 & 0.855 & 0.665 & $+1.72\;[+1.51,+1.92]$ & $34$\% \\
Swin / CIFAR-100 & 0.010 & 200 & 0.870 & 0.850 & $+6.47\;[+6.20,+6.75]$ & $65$\% \\
\addlinespace
DeiT / CIFAR-10 & 0 & 1000 & 0.040 & 0.001 & $-4.54\;[-4.90,-4.17]$ & -- \\
DeiT / CIFAR-10 & 0.0005 & 200 & 0.075 & 0.005 & $-4.01\;[-4.38,-3.63]$ & $-802$\% \\
DeiT / CIFAR-10 & 0.002 & 200 & 0.325 & 0.005 & $-2.63\;[-3.06,-2.20]$ & $-132$\% \\
DeiT / CIFAR-10 & 0.005 & 200 & 0.895 & 0.075 & $+0.27\;[-0.13,+0.67]$ & $5$\% \\
DeiT / CIFAR-10 & 0.010 & 200 & 1.000 & 0.470 & $+5.40\;[+5.05,+5.74]$ & $54$\% \\
\addlinespace
DeiT / CIFAR-100 & 0 & 1000 & 0.031 & 0.001 & $-4.05\;[-4.38,-3.73]$ & -- \\
DeiT / CIFAR-100 & 0.0005 & 200 & 0.135 & 0.010 & $-3.56\;[-3.83,-3.30]$ & $-712$\% \\
DeiT / CIFAR-100 & 0.002 & 200 & 0.345 & 0.130 & $-2.03\;[-2.46,-1.60]$ & $-102$\% \\
DeiT / CIFAR-100 & 0.005 & 200 & 0.510 & 0.415 & $+1.09\;[+0.85,+1.33]$ & $22$\% \\
DeiT / CIFAR-100 & 0.010 & 200 & 0.635 & 0.545 & $+5.69\;[+5.34,+6.04]$ & $57$\% \\
\bottomrule
\end{tabular}
\end{table}

For the profile probe, 20 repeats per setting at 0, 0.005 and 0.010 nats preceded
any weaker strength. A screen fixed before the audit required a positive lower 95\%
Monte Carlo limit for the mean expected-label log-loss gain at both positive
strengths, at least 25\% mean recovery at 0.010 nats, and no fitting failures or
optimiser warnings. The intervals describe simulation uncertainty given the fixed
caches and teachers.

% Generated by export_scalar_pilot.py from frozen profile recovery summaries.
\begin{table}[!htbp]
\centering\small
\caption{Profile expected-label binary log-loss recovery under the fixed
interaction teacher: 20 repeats per setting at each strength. Positive
gain is confidence-only loss minus Profile loss, in nats/example.
Percentages divide recovery by the exact oracle log-loss gain.}
\label{tab:profile-sensitivity}
\begin{tabular}{lrrrr}
\toprule
& \multicolumn{2}{c}{Oracle gain $0.005$} & \multicolumn{2}{c}{Oracle gain $0.010$} \\
Setting & Recovered & Fraction & Recovered & Fraction \\
\midrule
Swin / CIFAR-10 & $+0.001599$ & $32.0\%$ & $+0.005932$ & $59.3\%$ \\
Swin / CIFAR-100 & $+0.001307$ & $26.1\%$ & $+0.005170$ & $51.7\%$ \\
DeiT / CIFAR-10 & $+0.000285$ & $5.7\%$ & $+0.003290$ & $32.9\%$ \\
DeiT / CIFAR-100 & $-0.000540$ & $-10.8\%$ & $+0.002436$ & $24.4\%$ \\
\bottomrule
\end{tabular}
\end{table}

DeiT/CIFAR-100 fails the screen at 0.005 nats: mean gain $-0.000540$, 95\% Monte
Carlo interval $[-0.000719,-0.000360]$, and it recovers only 24.4\% at 0.010 nats.
The stop rule therefore applied, and the weaker profile strengths 0.0005 and 0.002
were not run; no teacher was tuned to improve recovery. The profile probe is thus
not uniformly sensitive to the specified 0.005-nat alternatives, and its failure to
improve on confidence cannot be read as a tight bound on conditional routing
information. This gate governed further simulation only; the separate continuation
rule for real correctness labels belongs to the correction probe of
App.~\ref{app:residual-probe}. The audits cover the binned diagnostic, the scalar
probe and the profile probe as specified here, and the complete-output probes of
App.~\ref{app:richer-output} with their original weight-decay grid; the
continuous-confidence models of App.~\ref{app:gap-robustness} and the
extended-grid learners of App.~\ref{app:richer-output} were not audited separately.

\subsection{Conditioning on the complete output}
\label{app:richer-output}
The primary estimand conditions on top-class confidence. A natural objection is
that confidence is a lossy summary of the output, so a profile probe could fail
against $c$ while still carrying information beyond the complete logit vector.
We therefore repeat the comparison with the full output as the conditioning set.

This is a complementary comparison, not a stricter one. Richer conditioning is
not monotone in conditional mutual information: $I(T;R\mid Z)$ can exceed or
fall below $I(T;R\mid c)$ even though $c$ is a function of $Z$, because a
routing variable and an output coordinate can be jointly predictive through an
interaction while neither is informative alone. Neither result implies the
other, and we report them separately.

\paragraph{Output representation.}
For raw logits $z\in\mathbb{R}^{K}$ we use
\begin{equation}
Z=\bigl[\,z-\textstyle\frac{1}{K}\sum_k z_k,\;\;
\textstyle\frac{1}{K}\sum_k z_k,\;\;
\operatorname{logit}(c),\;\;
\mathcal{H}(\operatorname{softmax}(z)),\;\;
p_{(1)}-p_{(2)}\,\bigr].
\label{eq:richer-output}
\end{equation}
Centred logits retain all softmax-probability information, and the mean-logit
coordinate restores the common shift, so the first $K+1$ columns are a lossless
recoding of finite raw logits; this is verified numerically for all 24 runs. The
remaining three columns are deterministic functions of the same logits, included
only to reduce learner burden. We do not divide each example's logits by its own
norm or standard deviation, which would discard scale information. Columns are
standardised using the training partition alone. The $K{=}10$ and $K{=}100$ runs
are fitted separately; no dataset is padded into another and no samples are
pooled across $K$.

\paragraph{Arms and selection.}
For each run we fit
\begin{equation}
\text{A}:\;\text{learner}([Z,\mathbf{0}_L]),\qquad
\text{B}:\;\text{learner}([Z,R]),\qquad
\text{C}:\;\text{learner}([Z,\pi(R)]),
\end{equation}
where $\pi$ permutes entire profiles, preserving layer correlations, drawn
independently inside each partition. All three arms share the outer folds of the
primary analysis, the same weight-decay grid $\{10^{-4},10^{-3},10^{-2}\}$,
the same two paired initialisation seeds, the same width of 32, the same epoch
budget, and the same inner-validation binary log-loss selection rule.
Seed-averaged validation predictions define each candidate, so selection scores
the ensemble that is actually reported. Each arm reverts to a shared
ridge-logistic reference fitted on $Z$ alone when inner validation does not
support the augmentation; that reference never sees $R$ and is identical across
arms. A small ridge-logistic triplet on the identical features provides an
inexpensive linear stability check. Nominal parameter matching through zero
padding does not equalise effective capacity, and we do not claim it does: the
comparison is the best pre-specified output-only procedure against its controlled
extension.

% Generated by scripts/revision_2026_09_25/render_revision.py from corrected_24run_2026-09-25/e2.
\begin{table}[!htbp]
\centering\small
\caption{Complete-output comparison on the real correctness labels: profile minus comparator in
binary log-loss ($10^{-3}$ nats); negative values favour the profile. \emph{Selected} applies the
selection rule, under which an arm falls back to the shared ridge output-only reference when inner
validation does not support its MLP; on the $K{=}100$ runs every selection fell back, so that column
is exactly zero there. \emph{MLP} and \emph{Ridge} compare like learners without selection. Pooled
rows as in Table~\ref{tab:repaired-primary}.}
\label{tab:e2-richer-output}
\setlength{\tabcolsep}{3pt}
\begin{tabular}{lrrrrr}
\toprule
& \multicolumn{3}{c}{Profile $-$ output-only} & \multicolumn{2}{c}{Profile $-$ permuted profile} \\
\cmidrule(lr){2-4}\cmidrule(lr){5-6}
Setting & Selected & MLP & Ridge & MLP & Ridge \\
\midrule
Swin / CIFAR-10 & $+0.26$ & $+1.40$ & $+0.07$ & $-0.82$ & $-0.29$ \\
Swin / CIFAR-100 & $+0.00$ & $+0.06$ & $+0.29$ & $-0.90$ & $-0.07$ \\
DeiT / CIFAR-10 & $+3.21$ & $+3.91$ & $+0.12$ & $-0.41$ & $-0.50$ \\
DeiT / CIFAR-100 & $+0.00$ & $+0.14$ & $+0.35$ & $-1.38$ & $-0.34$ \\
\midrule
$K{=}10$ runs ($n{=}12$) & $+1.73$ & $+2.66$ & $+0.09$ & $-0.62$ & $-0.40$ \\
$K{=}100$ runs ($n{=}12$) & $+0.00$ & $+0.10$ & $+0.32$ & $-1.14$ & $-0.20$ \\
\midrule
Pooled (24 runs) & $+0.87$ & $+1.38$ & $+0.21$ & $-0.88$ & $-0.30$ \\
95\% interval & $[-0.05,+2.46]$ & $[+0.05,+3.09]$ & $[-0.03,+0.46]$ & $[-1.52,-0.23]$ & $[-0.58,-0.03]$ \\
\bottomrule
\end{tabular}
\end{table}

\paragraph{Results.}
Table~\ref{tab:e2-richer-output} reports paired out-of-fold binary log-loss
differences, signed as routing-augmented minus comparator, so positive values
mean the routing-augmented arm is worse. The record-weighted profile minus
output-only difference is $+0.000867$ nats/example
($95\%$ CI $[-0.000046,+0.002457]$) under the selection rule,
$+0.001377$ ($[+0.000046,+0.003086]$) between the MLP arms, and
$+0.000207$ ($[-0.000029,+0.000459]$) between the ridge arms. Brier differences
have the same sign. Against its own permutation control the profile arm is
better: $-0.000880$ ($[-0.001521,-0.000229]$) for the MLP arms and
$-0.000298$ ($[-0.000580,-0.000029]$) for the ridge arms. The ridge triplet
reproduces both directions in all four settings individually.

Two of these are resolved at this precision and one is not. The
profile-minus-permuted-profile comparisons exclude zero in both learners. The
profile-minus-output-only comparisons do not: the selected and ridge intervals
both contain zero, so they preserve the ordering seen under confidence
conditioning without establishing it. The MLP-arm interval excludes zero, but
that arm is the one the selection rule most often rejects, so we do not treat
it as the headline. Conditioning on the complete output therefore leaves the
primary finding standing and adds no claim of its own about
$I(T;R\mid Z)$; because richer conditioning is not monotone in conditional
mutual information, neither comparison implies the other.

\paragraph{Why the selected contrast is zero at $K{=}100$.}
The fallback fired in $563$ of $720$ selections, and in all $360$
selections on the $K{=}100$ runs. There the out-of-fold predictions of arms A,
B and C are bit-identical to the reference, which we verified rather than
inferred, so the selected-model difference is exactly zero by construction and
the pooled selected column is carried entirely by the $12$ runs with $K{=}10$.
Decomposing the candidate risks on the same evaluation records explains the
reversion. Pooled over runs the routing-augmented network probe is worse than
its own output-only counterpart by $1.377\times10^{-3}$ nats and worse than the
ridge reference by $3.823\times10^{-3}$. Written as gains
($\mathcal{L}(\text{comparator})-\mathcal{L}(\text{routing MLP})$, positive
favouring routing), both pooled means are therefore negative, but the per-run signs
are not uniform: $5$ of $24$ runs have a positive gain on the first comparison and
$7$ of $24$ on the second. At $K{=}100$ the second gap is
$7.557\times10^{-3}$ and no run reverses it. Because the reference dominates
the observed candidate risks, reverting to it is consistent with predictive
risk minimisation on those splits. That is a statement about the candidates
actually fitted; it does not establish that every inner selection attained the
population optimum, and correct predictive-risk selection can coexist with low
sensitivity to a present signal. The corresponding statement about the
output-only arm is that the padded MLP trails the ridge reference by
$7.459\times10^{-3}$ at $K{=}100$ while leading it by $2.567\times10^{-3}$ at
$K{=}10$: the fitted MLPs in the two CIFAR-100 settings are not competitive
output-only learners here. We do not attribute that to output dimensionality
itself, which we have not ablated, and our results do not separate
function-class approximation error, finite-sample error, regularisation and
optimisation limits. These zeros are therefore not independent evidence that
routing adds nothing, and they do not bear on the ridge contrasts or the scalar
comparisons, which the rule does not touch. Choosing a candidate on the
evaluation set would close the gap arithmetically but would not yield a
deployable procedure, so we do not report one.

\paragraph{Weight-decay grid sensitivity.} Selection landed on the largest value
of the original grid in 462 of 720 fold-level selections. That grid was fixed
before any result of this experiment was seen. To check learner adequacy we
refitted all 24 runs post hoc with the grid extended to
$\{10^{-4},10^{-3},10^{-2},10^{-1},1\}$ for every MLP arm, keeping folds, seeds,
inner splits, the epoch rule, the reference and the selection and fallback rules
unchanged, and specified the extension before any outer-test result under it was
seen (Table~\ref{tab:e2-alpha-grid}). The output-only MLP now matches the reference
on average, fallbacks fall from 563 to 407, and the selected contrast shrinks to
$+0.35$ $[-0.18,+1.20]$; the MLP arms keep both orderings. The largest value, now 1,
is chosen in 636 selections, so stronger regularisation is still preferred and the
adequacy of these learners at the edge of the grid remains open. We did not extend
the grid again after seeing these results, and the extended grid was not included in
the positive control. In the two post-hoc recipe subsets of App.~\ref{app:probe-sanity}
the selected contrast on the original grid stays positive with an interval containing
zero ($+0.60$ $[-0.04,+2.03]$ and $+0.62$ $[-0.06,+2.04]$, record-weighted), and
both permuted-profile contrasts stay negative.

% Generated by scripts/revision_2026_09_25/e2_grid_latex.py.
\begin{table}[!htbp]
\centering\small
\caption{Post-hoc weight-decay grid sensitivity for the complete-output comparison (24 runs refitted; only the
MLP weight-decay grid changes, identically for all arms). Binary log-loss differences ($10^{-3}$ nats; negative
favours the first-named predictor) with pooled 95\% hierarchical-bootstrap intervals. Ridge rows do not depend
on the grid.}
\label{tab:e2-alpha-grid}
\setlength{\tabcolsep}{4pt}
\begin{tabular}{lcc}
\toprule
& Grid $\{10^{-4},10^{-3},10^{-2}\}$ & Grid $\{10^{-4},\ldots,1\}$ \\
\midrule
Profile $-$ output-only, selected & $+0.87\;[-0.05,+2.46]$ & $+0.35\;[-0.18,+1.20]$ \\
Profile $-$ output-only, MLP & $+1.38\;[+0.05,+3.09]$ & $+0.75\;[+0.25,+1.46]$ \\
Profile $-$ output-only, ridge & $+0.21\;[-0.03,+0.46]$ & $+0.21\;[-0.03,+0.46]$ \\
Profile $-$ permuted, MLP & $-0.88\;[-1.52,-0.23]$ & $-0.78\;[-1.19,-0.39]$ \\
Profile $-$ permuted, ridge & $-0.30\;[-0.58,-0.03]$ & $-0.30\;[-0.58,-0.03]$ \\
Output-only MLP $-$ reference & $+2.45\;[-2.77,+7.68]$ & $-0.14\;[-3.46,+3.02]$ \\
\midrule
Fallbacks to reference (of 720) & 563 & 407 \\
\quad on $K{=}100$ runs (of 360) & 360 & 337 \\
Largest grid value selected (of 720) & 462 & 636 \\
\bottomrule
\end{tabular}
\end{table}

\paragraph{Positive control.}
Before using the injected-signal generator here we checked what it probes.
Under its law $Q$, the anchor supplies $Z$ and a donor drawn from the anchor's
confidence block supplies $R$, so
$Q(Z,R,T^*\mid c)=Q(Z\mid c)\,Q(R\mid c)\,Q(T^*\mid c,R)$: donors are drawn
from the same block for every anchor sharing a confidence value, and we verified
that the conditional label probability varies by at most $6\times10^{-17}$
within each $(c,R)$ cell. Computing
$q_Q(z)=\mathbb{E}_Q[T^*\mid Z=z]$ and
$\eta_Q(z,r)=\mathbb{E}_Q[T^*\mid Z=z,R=r]$ directly from the generation
probabilities gives
$G_Z=\mathbb{E}_Q\,D_{\rm KL}(\mathrm{Bern}(\eta_Q)\Vert\mathrm{Bern}(q_Q))$
equal to the $c$-conditioned oracle gain to eight decimals in all four caches.
We do not rest this on confidence being injective: two caches contain tied
confidence values with positive empirical mass, and we checked that those
anchors are exact duplicate records, so they share $Z$, $q_0$ and the solved
intercept. All of this is a property of $Q$, not of the real-data law.

Because $q_Q(z)=q_0(c)$, no measurable function of $Z$ alone can improve on
$q_0$ under $Q$; that is analytic. Separately, the fitted output-only arms score
\emph{worse} than $q_0$ in every cell, with Monte Carlo upper limits below zero
at both strengths. That second statement is a negative-control result about
those particular estimators and their finite-sample excess risk, and it is not
evidence about all $Z$-only functions. The gate record is retained in the
supplementary material.

Across $240$ replicates ($4$ settings $\times$ $\{0,0.005,0.010\}$ nats
$\times$ $20$ repeats, reusing the seeds of the confidence-conditioned injected-signal
experiment in App.~\ref{app:profile-sensitivity}, with no failures or
optimiser warnings), decomposing the candidate risks separates three quantities
that a single contrast conflates. Write $\mathcal{L}$ for expected-label
log-loss. The within-family gain
$\mathcal{L}(\text{MLP}_Z)-\mathcal{L}(\text{MLP}_{ZR})$ is positive in
$20/20$ replicates in seven of the eight positive-strength cells, so the
network probe does recover the injected effect from routing. The gain against
the deployed reference, $\mathcal{L}(\text{REF})-\mathcal{L}(\text{MLP}_{ZR})$,
is positive in $20/20$ replicates on the $K{=}10$ settings at $0.010$ nats but
in $0/20$ on both $K{=}100$ settings. The quantity that decides this is the output-only probe's own deficit against
the reference, $D_{\rm base}=\mathcal{L}(\text{MLP}_Z)-\mathcal{L}(\text{REF})
=G_{\rm within}-G_{\rm vs\,REF}$, evaluated on the same replicates. At $0.010$
nats $D_{\rm base}$ is $7.30\times10^{-3}$ on Swin/CIFAR-100 and
$6.24\times10^{-3}$ on DeiT/CIFAR-100, while the within-family gain closes
$21.8\%$ and $57.4\%$ of it. The routing gain is real and, in one setting,
closes most of the gap; it does not close all of it, so the augmented candidate
remains behind the reference and the selected contrast follows that second
comparison. The residual gap after augmentation is not the original gap, and we
do not describe the two as differing by an order of magnitude. Where the
augmented candidate is not the best of those fitted, deploying it would raise
held-out loss on those splits, so the reversion is consistent with predictive
risk minimisation. We report this as a limit of the MLPs fitted in these two
CIFAR-100 settings rather than of output dimensionality, which we have not
ablated, and we do not weaken the reference in order to produce a positive
difference. Correct predictive-risk selection and low sensitivity to a present
signal are compatible, and this control does not show that the injected signal
was not missed.

Two limits on what this control licenses. Its teacher acts through the donor
profile, which is conditionally independent of the anchor output given the
confidence block, so it exercises recovery of profile signal orthogonal to $Z$
given $c$ and says nothing about interactions between $Z$ and $R$. And passing
one teacher does not validate all interaction signals, just as failing one does
not invalidate all non-interaction signals. A recovered-gain fraction is not a
detection probability, and neither is statistical power at a fixed test level;
we report all three separately in the supplementary material.

\section{Supporting propositions}
\label{app:theory}

These supporting results connect the population target to the empirical
tests. Propositions~1 and~2 characterise the value of incremental correctness
information and the consequence of a routing-defined subgroup gap. The
finite-sample discussion separates binned from continuous conditioning and
accounts for multiplicity. Standard fact~3 and Proposition~4 describe
estimator recovery and bandwidth attenuation. All are standard or supporting
results, not finite-sample guarantees or dominance claims for AR-CondCal.

\textbf{Setup.}
Let $(X, Y)$ be a classification input--label pair from the
population distribution, $\hat{y}(X) = \arg\max_k p_k(X)$ the argmax
prediction, and $T := \mathbb{1}[\hat{y}(X) = Y] \in \{0, 1\}$ the
binary correctness indicator. Let $c(X) = \max_k p_k(X)$ be the
top-class confidence and $r(X)$ a possibly-vector-valued,
routing-derived feature (in AR-CondCal,
$r = r_{\mathrm{std}}$, the depth standard deviation of per-layer routing
entropy). Define the oracle conditional correctness functions
\begin{equation}
\eta_c(c) := \mathbb{E}[T \mid c(X) = c],
\qquad
\eta_{c,r}(c, r) := \mathbb{E}[T \mid c(X) = c,\, r(X) = r].
\end{equation}
We assume $T$ has finite second moment throughout.

\paragraph{Population value of routing.}
The relevant population question is not whether routing varies, but
whether it changes conditional correctness after confidence is fixed.
The next identity gives the exact oracle value of that information.

\textbf{Proposition 1 (oracle proper-score gains).}
\emph{For binary correctness, the reduction in oracle Brier risk from
adding $r$ to $c$ equals the $L^2$ distance between the two conditional
means:}
\begin{equation}
\label{eq:explained-variance}
\mathbb{E}\bigl[(T - \eta_c(c(X)))^2\bigr]
\;-\;
\mathbb{E}\bigl[(T - \eta_{c,r}(c(X), r(X)))^2\bigr]
\;=\;
\mathbb{E}\bigl[(\eta_{c,r}(c(X), r(X)) - \eta_c(c(X)))^2\bigr] \;\ge\; 0.
\end{equation}
\emph{The corresponding reduction in oracle binary log-loss is}
\begin{equation}
\label{eq:logloss-information}
R_{\log}(c)-R_{\log}(c,r)
=H(T\mid c)-H(T\mid c,r)
=I(T;r\mid c)\;\geq\;0.
\end{equation}
\emph{For either score, the oracle gain is strictly positive iff
$\eta_{c,r}(c,r) \neq \eta_c(c)$ on a positive-mass set, and is
zero iff $r$ carries no additional information about $T$ given
$c$ almost surely.}

\emph{Proof.} Writing $A := \eta_{c,r}(c(X), r(X))$ and
$B := \eta_c(c(X))$, both $A$ and $B$ are conditional expectations
of $T$ with respect to $\sigma$-algebras
$\sigma(c, r) \supseteq \sigma(c)$, so $A - B$ is
$\sigma(c, r)$-measurable and $\mathbb{E}[T - A \mid \sigma(c, r)]
= 0$. Decomposing $T - B = (T - A) + (A - B)$ and using this
orthogonality gives
$\mathbb{E}[(T - B)^2] = \mathbb{E}[(T - A)^2] + \mathbb{E}[(A - B)^2]$.
Rearranging gives~\eqref{eq:explained-variance}. For log-loss, the
oracle Bernoulli forecast under an information set is its conditional
mean, so its expected log-loss is the corresponding conditional
entropy. Their difference is the definition of conditional mutual
information, giving~\eqref{eq:logloss-information}.~\hfill$\square$

\textbf{Remark.}
These are standard oracle identities specialised to the AR calibration
question, not new information-theoretic results. They align the
population question with the Brier and log-loss scores used in the
direct-correctness audit. Equation~\eqref{eq:explained-variance}
motivates comparing routing summaries by the incremental correctness
variation they could in principle explain. The feature ablation in
Tab.~\ref{tab:feature_ablation} is only a finite-sample ECE-based probe
of this question and is not an estimator of the right-hand side of
Eq.~\eqref{eq:explained-variance}. The identity does \emph{not} imply
that any particular finite-sample kernel calibrator achieves
this gain, nor that aggregating over a binned ECE metric
preserves it.

\textbf{Excess-risk decomposition.} For fitted predictors $f_c$ and $f_{c,R}$
evaluated on the same distribution, write $e_c=L(f_c)-R_{\log}(c)$ and
$e_{c,R}=L(f_{c,R})-R_{\log}(c,R)$ for their excess log-loss risks above the
respective oracles. Subtracting and using Eq.~\eqref{eq:logloss-information}
gives Eq.~\eqref{eq:excess-risk},
$L(f_c)-L(f_{c,R})=I(T;R\mid c)+e_c-e_{c,R}$. A fitted contrast therefore
identifies the information term only if the excess risks are known; a negative
contrast is compatible with $I(T;R\mid c)>0$ whenever $e_{c,R}>e_c+I(T;R\mid c)$.
The sensitivity audit measures this gap on specified alternatives.

\paragraph{Finite-sample detection.}
Even when the condition in Proposition~1 holds, a finite sample may not
resolve it. Proposition~2 motivates subgroup comparison, after which we
state what its binned empirical counterpart does and does not identify.

\textbf{Proposition 2 (confidence-only blind spot at routing-defined
subgroups).}
\emph{Let $G : \mathcal{X} \to \{1, \dots, K\}$ be any measurable
subgroup labelling derived from $r(X)$ (e.g., tertiles of
$r_{\mathrm{agg}}$ or $r_{\mathrm{std}}$), and define the
within-subgroup conditional accuracy at confidence $c$ as
$\eta_g(c) := \mathbb{E}[T \mid c(X) = c,\, G(X) = g]$.
All subgroup maxima below range over groups with positive conditional
probability at $c$. Let
$\widehat{\pi}_\circ : [0, 1] \to [0, 1]$ be any post-hoc
calibrator whose output is a measurable function of confidence
alone, and consider its worst-subgroup mismatch at confidence
$c$,}
\begin{equation}
\Delta_\circ(c) \;:=\; \max_{g}\, |\widehat{\pi}_\circ(c) - \eta_g(c)|.
\end{equation}
\emph{For almost every $c$ at which the chosen regular conditional
versions are defined and two subgroup conditional accuracies
differ, i.e.\ $\eta_g(c) \neq \eta_{g'}(c)$ for some $g, g'$,}
\begin{equation}
\Delta_\circ(c)
\;\ge\;
\tfrac{1}{2}\,
\max_{g, g'}\, |\eta_g(c) - \eta_{g'}(c)|
\;>\; 0,
\end{equation}
\emph{with equality achievable only at the midpoint of the extreme
subgroup accuracies. Consequently, no confidence-only post-hoc
calibrator can drive the worst-subgroup mismatch to zero at almost
every confidence value where subgroup accuracies separate.}

\emph{Proof.} For any scalar $y$ and any reals $a \neq b$,
$\max(|y - a|, |y - b|) \ge \tfrac{1}{2}|a - b|$, with equality
iff $y = (a + b)/2$. Apply this with $a = \eta_g(c)$,
$b = \eta_{g'}(c)$ for the pair achieving the maximum spread and
$y = \widehat{\pi}_\circ(c)$.~\hfill$\square$

\textbf{Remark.}
At confidence values shared by two routing subgroups, a single
confidence-only prediction cannot match different conditional accuracies.
This motivates the subgroup diagnostic in Sec.~\ref{sec:rq1}.
The empirical statistic compares low and high tertiles within confidence
bins and then maximises over bins. It is not a direct estimate of the
pointwise population bound, which conditions on continuous confidence
and ranges over subgroups. The proposition supplies a reason to test for
subgroup differences; their magnitude and reproducibility must be
established empirically.

The empirical max-over-bins statistic replaces continuous $c$ by a
finite partition and then selects an extreme across bins. It therefore
targets a binned approximation, can miss within-bin confounding, and has
high variance where support is sparse. This is why our protocol varies
the bin count, simulates Type-I error on the observed support, and adds
continuous-confidence logistic correctness prediction and local-caliper checks. Multiplicity is a
separate issue: if $R=\sum_{j=1}^{m}\mathbf{1}\{p_j\leq\alpha\}$ under
a global null with valid marginal $p$-values, then
\begin{equation}
\mathbb{E}[R]=\sum_{j=1}^{m}\Pr(p_j\leq\alpha)\leq m\alpha,
\qquad
\Pr\!\left(\min_j p_j\leq\alpha/m\right)\leq\alpha.
\label{eq:multiplicity-accounting}
\end{equation}
With $m=30$ and $\alpha=0.05$ the first relation gives an expected count of 1.5
nominal rejections under exactly uniform null $p$-values; it fixes the mean, not the
distribution, of $R$ when the tests are dependent, so observing a nominal hit and a
borderline result is a reference point rather than a test of the joint null. The
second relation is the Bonferroni family-wise bound and requires no independence.
Neither proves the null, and the power of the binned test against specified
alternatives is reported in Table~\ref{tab:scalar-sensitivity}.

\paragraph{Estimator recovery.}
Population existence and finite-sample evidence do not guarantee that a
chosen projection, bandwidth, and estimator produce held-out benefit.
The standard asymptotic fact below establishes only eventual recovery
under regularity conditions; Proposition~4 isolates a concrete
finite-bandwidth failure mode.

\textbf{Standard fact 3 (Nadaraya--Watson consistency).}
\emph{For i.i.d.\ calibration triples, under standard pointwise
kernel-regression regularity conditions
(a bounded kernel $K$ that integrates to one; bandwidth $h_n \to 0$
with $n h_n^d \to \infty$ for feature dimension $d=2$; continuity of
$\eta_{c,r}$; and a positive, continuous joint density of $(c,r)$ at
the interior point of interest), the
Nadaraya--Watson estimator}
\begin{equation}
\hat{\eta}_{c,r}(c, r)
= \frac{\sum_{i=1}^{n} K_{h_n}(c - c_i,\, r - r_i)\; T_i}
       {\sum_{i=1}^{n} K_{h_n}(c - c_i,\, r - r_i)}
\end{equation}
\emph{satisfies $\hat{\eta}_{c,r}(c, r) \to \eta_{c,r}(c, r)$ in
probability as $n \to \infty$ at interior points of the support}
\citep{nadaraya1964estimating,watson1964smooth,gyorfi2002distribution}.

\textbf{Remark.}
This asymptotic statement explains why kernel regression can target
$\eta_{c,r}$ under appropriate sampling conditions. It supplies no
finite-sample guarantee for the calibration sizes in this study;
the size and bandwidth analyses examine those practical limitations
(Apps.~\ref{app:calsize} and~\ref{app:bw-sensitivity}).

\textbf{Proposition 4 (stylised bandwidth-induced subgroup-gap
shrinkage; two-group, $1$-D Gaussian Nadaraya--Watson).}
\emph{Consider two equal-mass routing subgroups located at
$r = -d/2$ and $r = +d/2$ with within-subgroup conditional means
$f_-$ and $f_+$ at a fixed confidence $c$. Let
$\widehat{s}(r)$ denote the $1$-D Gaussian Nadaraya--Watson
estimate at routing-feature value $r$ and bandwidth $h$,}
\begin{equation}
\widehat{s}(r)
\;=\;
\frac{\,e^{-(r + d/2)^2 / (2h^2)}\, f_-
   \;+\; e^{-(r - d/2)^2 / (2h^2)}\, f_+\,}
     {\,e^{-(r + d/2)^2 / (2h^2)}
   \;+\; e^{-(r - d/2)^2 / (2h^2)}\,}.
\end{equation}
\emph{Then the smoothed subgroup gap evaluated at the group centres
satisfies}
\begin{equation}
\label{eq:bandwidth-shrinkage}
\widehat{s}(-d/2) \;-\; \widehat{s}(+d/2)
\;=\; (f_- - f_+)\,\tanh\!\bigl(d^2 / (4h^2)\bigr).
\end{equation}
\emph{The shrinkage factor $\tanh(d^2 / (4h^2))$ tends to $0$ as
$h / d \to \infty$ and to $1$ as $h / d \to 0$.}

\emph{Proof.} Substituting $r = \mp d/2$ and writing
$w := e^{-d^2 / (2h^2)}$,
$\widehat{s}(-d/2) = (f_- + w f_+) / (1 + w)$ and
$\widehat{s}(+d/2) = (w f_- + f_+) / (1 + w)$, so~\,
$\widehat{s}(-d/2) - \widehat{s}(+d/2)
 = (f_- - f_+)\,(1 - w)/(1 + w)$.
The identity
$(1 - e^{-x})/(1 + e^{-x}) = \tanh(x/2)$ with
$x = d^2/(2h^2)$ gives the factor
$\tanh(d^2/(4h^2))$.~\hfill$\square$

\textbf{Remark.}
This is a stylised two-regime calculation, not a finite-sample
theorem for the actual $2$-D AR-CondCal estimator. It illustrates a
possible bandwidth-versus-spacing failure mode: when the
ratio of kernel bandwidth to subgroup spacing, $h/d$, grows, the
retained fraction of the population subgroup gap tends to zero. We do not map this stylised factor to the
empirical cells without directly estimating both $d$ and $h$; the
proposition is explanatory rather than an empirical attribution.

\textbf{Scope.}
The identities concern population Brier and log-loss risk, whereas ECE is a
binned, non-$L^2$ functional and the probes are finite-sample estimators.
Their empirical ordering can depend on feature range, bandwidth, and sample
size, as the benchmark and sensitivity analyses show. The propositions make
no claim about baseline dominance, classification accuracy, adversarial or
out-of-distribution robustness, or safety. Their role is to define the target,
motivate the subgroup diagnostic, justify kernel regression pointwise, and
illustrate finite-bandwidth attenuation.

\section{Internal comparator: MEC}
\label{app:internal-baselines}

\textbf{MEC (Margin--Entropy Calibration)} is a logits-only internal
comparator retained in Tab.~\ref{tab:calibration_block_ar}; it is not a
claimed contribution or an externally published baseline. Its inputs
are the top-1/top-2 logit margin, predictive entropy, and optionally
the predicted class for a class-specific offset. It consumes no AR
routing weights or routing-entropy features.

MEC first fits a global temperature-scaling anchor. A class offset is
enabled only when the calibration split exhibits sufficient
between-predicted-class variation. It then forms quantile bins for the
margin and predictive entropy and, when either feature has sufficient
Spearman association with the residual calibration error, fits
additive per-bin log-temperature offsets by L-BFGS with $L^2$
regularisation. Calibration applies the resulting sample-specific
temperature to the full logit vector. We make no monotonicity claim. The
supplementary README maps the implementation's class and result keys to MEC.

\FloatBarrier
\section{Cross-architecture and multi-seed evidence}
\label{app:cross-arch-multiseed}

\textbf{ViT-B/16 runs.} The six ViT-B/16 runs provide only
cross-architecture and cross-dataset scope context. Every value in
Tab.~\ref{tab:vit-pilot-compact} is from a single seed and is descriptive only;
none enters a replicated architecture claim or the 24-run pooled analyses. The
full calibration leaderboard and feature-ablation history are omitted because
single-run method ordering cannot establish stability.

\begin{table}[!ht]
\centering
\small
\setlength{\tabcolsep}{7pt}
\caption{Compact ViT-B/16 routing diagnostic. Every row is one seed and is
descriptive only. Max gap and $p$ use the same fixed $r_{\mathrm{agg}}$
bin-conditional diagnostic as the 30-test family.}
\label{tab:vit-pilot-compact}
\begin{tabular}{llcc}
\toprule
Dataset & Variant & Max gap & Permutation $p$ \\
\midrule
CIFAR-10 & Block-AR & $0.1255$ & $0.421$ \\
CIFAR-10 & Full-AR  & $0.2222$ & $0.511$ \\
CIFAR-100 & Block-AR & $0.0936$ & $0.360$ \\
CIFAR-100 & Full-AR  & $0.0672$ & $0.827$ \\
Tiny-ImageNet & Block-AR & $0.0898$ & $0.449$ \\
Tiny-ImageNet & Full-AR  & $0.0984$ & $0.318$ \\
\bottomrule
\end{tabular}
\end{table}

\textbf{Other training runs.} Two further Swin-Tiny/CIFAR-10 Block-AR checkpoints
(seeds 0 and 2) were trained with the implementation's default block setting: their
resolved configurations set no block size. Seed 0 did not train normally (test accuracy
0.74). Neither is pooled with the $b=2$ runs or enters any analysis; their checkpoint and
configuration hashes are listed in the supplementary material.

\subsection{The 30-test diagnostic family}
\label{app:replication-plan}

Table~\ref{tab:ar-sweep} aggregates the 30 tests by setting and variant. Swin-Tiny
and DeiT-Small rows follow the replicated three-seed protocol; ViT-B/16 rows are
single-seed runs that belong to the multiplicity family but support no replicated
claim. Restricting the family to the 24 replicated runs (threshold $0.05/24$) leaves
the multiplicity conclusion unchanged. Per-run values are in the supplementary
material.

\begin{table}[!ht]
\centering
\caption{The 30-test diagnostic family aggregated by setting and variant.
Replicated rows report mean $\pm$ standard deviation of the maximum gap over three
seeds, the smallest seed-level permutation $p$ (5,000 permutations) and the number
of seeds with $p<0.05$; ViT-B/16 rows ($\ast$) are single runs.}
\label{tab:ar-sweep}
\footnotesize
\setlength{\tabcolsep}{5pt}
\begin{tabular}{llccc}
\toprule
\textbf{Setting} & \textbf{Variant} & $n$ & \textbf{Max gap} & \textbf{Min $p$, $p<0.05$} \\
\midrule
\multirow{2}{*}{Sw-T / C-10}
  & Block-AR ($b=2$) & $3$ & $0.1927\,{\scriptstyle\pm 0.1334}$ & $0.0382$, $1/3$ \\
  & Full-AR          & $3$ & $0.1336\,{\scriptstyle\pm 0.0418}$ & $0.5469$, $0/3$ \\
\addlinespace
\multirow{2}{*}{DeiT-S / C-10}
  & Block-AR ($b=1$) & $3$ & $0.1222\,{\scriptstyle\pm 0.0276}$ & $0.2647$, $0/3$ \\
  & Full-AR          & $3$ & $0.1152\,{\scriptstyle\pm 0.0154}$ & $0.4951$, $0/3$ \\
\addlinespace
\multirow{2}{*}{DeiT-S / C-100}
  & Block-AR ($b=1$) & $3$ & $0.0941\,{\scriptstyle\pm 0.0208}$ & $0.4519$, $0/3$ \\
  & Full-AR          & $3$ & $0.0912\,{\scriptstyle\pm 0.0285}$ & $0.0494^{\S}$, $1/3^{\S}$ \\
\addlinespace
\multirow{2}{*}{Sw-T / C-100}
  & Block-AR ($b=2$) & $3$ & $0.1092\,{\scriptstyle\pm 0.0136}$ & $0.0934$, $0/3$ \\
  & Full-AR          & $3$ & $0.0902\,{\scriptstyle\pm 0.0334}$ & $0.1442$, $0/3$ \\
\addlinespace
ViT-B/$16$ / C-100 & Block-AR$^{\ast}$ & $1$ & $0.0936$ & $0.360$, $0/1$ \\
ViT-B/$16$ / C-100 & Full-AR$^{\ast}$  & $1$ & $0.0672$ & $0.827$, $0/1$ \\
ViT-B/$16$ / C-10  & Block-AR$^{\ast}$ & $1$ & $0.1255$ & $0.421$, $0/1$ \\
ViT-B/$16$ / C-10  & Full-AR$^{\ast}$  & $1$ & $0.2222$ & $0.511$, $0/1$ \\
ViT-B/$16$ / T-IN  & Block-AR$^{\ast}$ & $1$ & $0.0898$ & $0.449$, $0/1$ \\
ViT-B/$16$ / T-IN  & Full-AR$^{\ast}$  & $1$ & $0.0984$ & $0.318$, $0/1$ \\
\bottomrule
\end{tabular}

\smallskip
\noindent\scriptsize Sw-T = Swin-Tiny, DeiT-S = DeiT-Small, C-10/C-100 =
CIFAR-10/100, T-IN = Tiny-ImageNet. $^{\ast}$Single-seed run.
$^{\S}$Borderline run (seed 1): $p=0.0494$ with 5,000 permutations, and 0.0479--0.0503
in five recomputations with 50,000 permutations (App.~\ref{app:gap-robustness}); it is
counted below $0.05$ only at the 5,000-permutation value, and every value lies far
above $0.05/30$.
\end{table}

\clearpage
\section{Routing-feature geometry and layer-wise structure}
\label{app:mechanism}

\begin{figure}[!htbp]
  \centering
  \includegraphics[width=\linewidth]{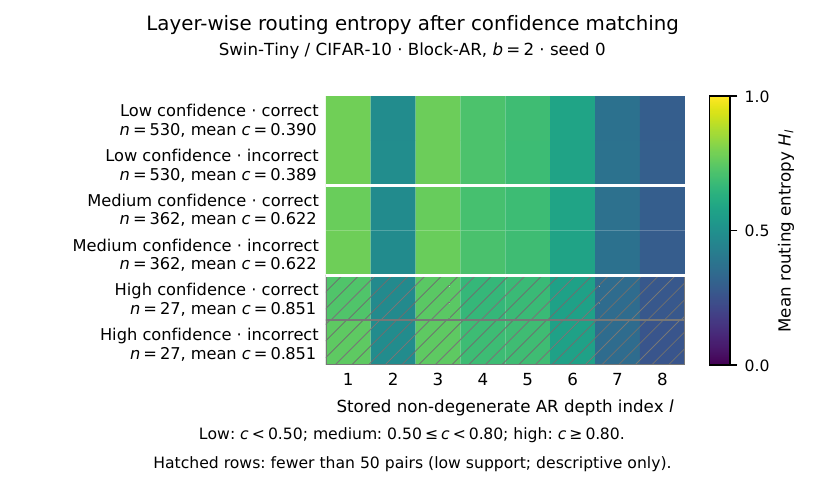}
  \caption{\textbf{Layer-wise routing entropy after confidence matching.}
  For a representative Swin-Tiny/CIFAR-10 Block-AR run ($b=2$, seed $0$),
  correct and incorrect predictions are matched $1{:}1$ without replacement
  within fixed confidence bands with $|\Delta c|\leq0.005$. Columns show
  the stored non-degenerate AR layers and colours show mean normalised
  routing entropy on a shared $[0,1]$ scale. Row labels report sample counts
  and mean confidence; the low, medium, and high bands contain $530$, $362$,
  and $27$ pairs, respectively. Correct and incorrect profiles are visually
  similar in the better-supported bands. High-confidence rows contain fewer
  than $50$ pairs and are hatched and labelled as low support. This
  visualisation is descriptive and is not a conditional-independence test.}
  \label{fig:layerwise-entropy-profiles}
\end{figure}

\begin{figure}[!ht]
  \centering
  \includegraphics[width=0.9\linewidth]{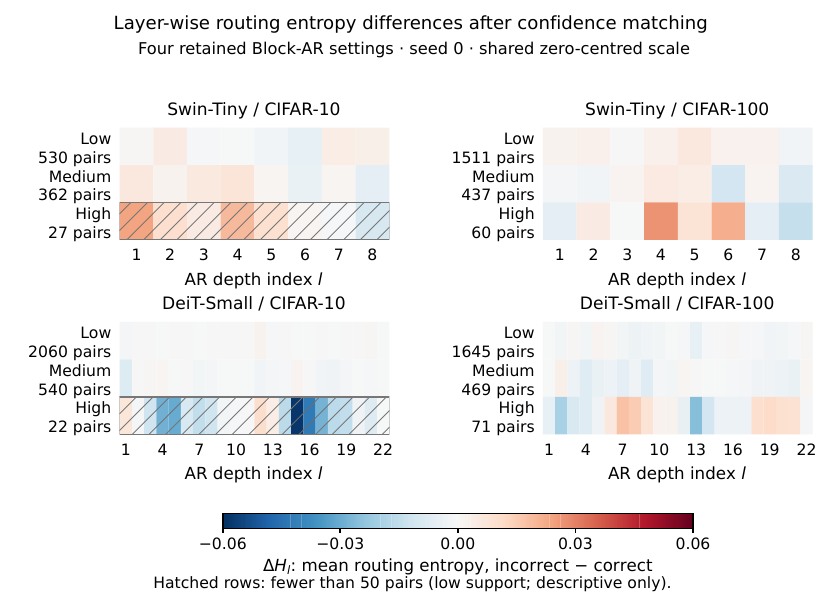}
  \caption{\textbf{Layer-wise routing entropy differences after confidence
  matching.} For the seed-$0$ Block-AR runs of the four
  replicated backbone--dataset settings, correct and incorrect predictions
  are matched $1{:}1$ without replacement within fixed confidence bands
  ($c<0.50$, $0.50\leq c<0.80$, $c\geq0.80$; $|\Delta c|\leq0.005$).
  Each cell shows the difference in mean per-layer routing entropy,
  incorrect minus correct, using a shared zero-centred colour scale:
  positive values indicate higher routing entropy for incorrect predictions.
  The better-supported low- and medium-confidence groups show small,
  setting-dependent differences, with no visually common depth-localised
  pattern across the four settings. Rows with fewer than $50$ pairs are
  hatched and labelled as low support. Depth indices follow each
  architecture's stored non-degenerate AR layers. This figure describes
  feature geometry in the matched subsets and is not a
  conditional-independence test; the replicated diagnostics in
  Sec.~\ref{sec:rq1} provide the inferential evidence.}
  \label{fig:layerwise-entropy-differences}
\end{figure}

Figures~\ref{fig:layerwise-entropy-profiles} and
\ref{fig:layerwise-entropy-differences} visualise the cached routing-entropy
profile $R=(H_1,\ldots,H_L)$ after close confidence matching. We use the
seed-$0$ Block-AR run of each backbone--dataset setting,
including the $b=2$ Swin-Tiny/CIFAR-10 retrain. Confidence $c$ is the maximum
softmax probability from the raw logits. Starting from each run's
$10{,}000$ test examples, we form fixed bands $c<0.50$,
$0.50\leq c<0.80$, and $c\geq0.80$. Within each band, incorrect examples
in ascending confidence order are paired with their nearest unused correct
example, retaining pairs only when $|\Delta c|\leq0.005$; ties use cache
order. Matching uses no routing values. The plots describe this matched
overlap subset, not the complete test population or exact conditioning on
$c$. The Block-AR profiles have $L=8$ non-degenerate layers for Swin-Tiny ($b=2$)
and $L=22$ for DeiT-Small ($b=1$); layer indices are architecture-specific.

\subsection{Scalar routing-feature geometry and ablation}

\begin{figure}[h]
  \centering
  \includegraphics[width=\linewidth]{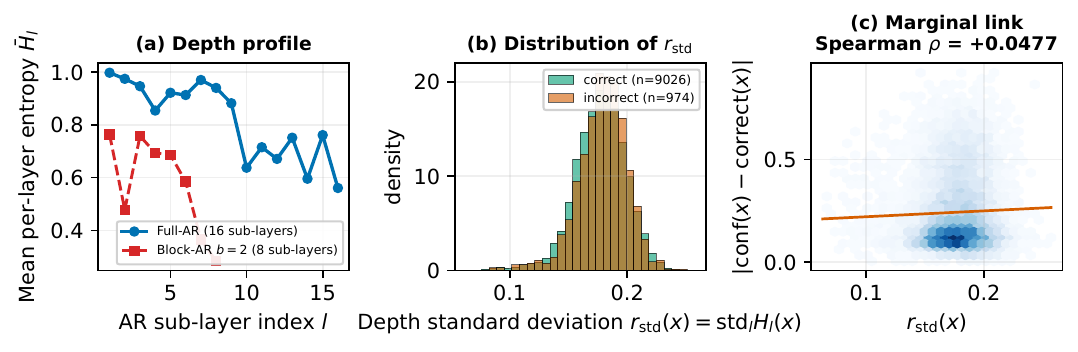}
  \caption{\textbf{Routing-feature geometry on the Swin-Tiny/CIFAR-10 Block-AR
  ($b{=}2$) seed-0 run} (accuracy $0.9026$; $9{,}026$ correct and $974$ incorrect of
  $10{,}000$), with the Full-AR seed-0 run for the depth comparison in \emph{(a)}.
  Every quantity is computed from the routing caches of these runs. \emph{(a)} Mean per-layer
  entropy at $8$ versus $16$ AR sub-layers. \emph{(b)} Raw depth standard
  deviation $r_{\mathrm{std}}$, not rescaled, for correct versus incorrect
  predictions. \emph{(c)} Marginal rank association between
  $r_{\mathrm{std}}$ and the per-sample absolute residual
  $|\mathrm{conf}-\mathrm{correct}|$. That residual is not conditional
  correctness and not a population calibration error, and this panel is not a
  test of conditional independence.}
  \label{fig:mechanism}
\end{figure}

Figure~\ref{fig:mechanism} examines the routing-feature geometry
on the Swin-Tiny Block-AR ($b{=}2$) CIFAR-10 seed-$0$ run analysed in
Tab.~\ref{tab:phenomenon}, and reports the marginal rank association between the chosen
second feature and the per-sample absolute residual $|\mathrm{conf}-\mathrm{correct}|$.

\textbf{(a) Depth profile.} Mean per-layer routing entropy $\bar H_l$ is not
interchangeable between the two variants on this pair: Block-AR $b{=}2$ has
$8$ AR sub-layers with cross-depth standard deviation of $\bar H_l$ equal to
$0.171$, while Full-AR has $16$ sub-layers and $0.142$. The two curves are
therefore compared at different depths and are described, not equated. This
panel says nothing about equality of the population routing distributions.

\textbf{(b) Sample-level distribution.} The depth standard deviation
$r_{\mathrm{std}}(x)=\mathrm{std}_l H_l(x)$ is plotted raw, with no rescaling,
over its observed range $[0.063,0.257]$. Its distributions for correct and
incorrect predictions overlap almost completely: the conditional means are
$0.1769$ and $0.1790$. A one-dimensional threshold on $r_{\mathrm{std}}(x)$
could not separate the two.

\textbf{(c) Marginal rank association.} The Spearman correlation between
$r_{\mathrm{std}}(x)$ and the per-sample absolute residual
$|\mathrm{conf}(x)-\mathrm{correct}(x)|$ is $\rho=+0.0477$
($p=1.8\times10^{-6}$, $n=10{,}000$) in this cell. The quantity correlated
with is that per-sample residual, which is neither conditional correctness
given confidence nor a population calibration error; no statement about
either follows from this panel. The statistic does not test conditional
independence given confidence, and the conditional analyses in
\S\ref{sec:rq1}--\S\ref{sec:probe} are what address that question.

\begin{table}[!ht]
\centering
\setlength{\tabcolsep}{3.5pt}
\small
\caption{\textbf{Feature ablation across auxiliary $1$-D summaries and controls} (Swin-Tiny seed-$0$). Confidence-only uses the genuinely $1$-D $\mathrm{NW}(c)$ control; every auxiliary-feature row uses the corresponding $2$-D $\mathrm{NW}(c,f)$ estimator under the same split and kernel convention. Confidence and predictive entropy are non-routing controls; $1-r_{\mathrm{agg}}$ is redundant with $r_{\mathrm{agg}}$. Across these point estimates, no routing-derived scalar consistently outperforms the controls, and the best routing summary changes across substrates. The table is descriptive: it neither establishes cross-seed equivalence nor rules out other scalar projections.}
\label{tab:feature_ablation}
\begin{tabular}{p{8.0cm}rr}
\toprule
Feature & Block-AR ECE $\downarrow$ & Full-AR ECE $\downarrow$ \\
\midrule
\multicolumn{3}{l}{\textbf{CIFAR-10} (Swin-Tiny seed-$0$, $b\!=\!2$ for Block-AR)} \\
\midrule
Confidence only & 0.0131 & 0.0232 \\
Predictive entropy & 0.0092 & 0.0199 \\
Aggregate routing entropy $r_{\mathrm{agg}}$ & 0.0099 & 0.0241 \\
Last-layer routing entropy $H_L$ & 0.0075 & 0.0191 \\
Routing concentration $1\!-\!r_{\mathrm{agg}}$ & 0.0099 & 0.0241 \\
Routing entropy $\times$ confidence & 0.0136 & 0.0207 \\
Depth standard deviation $r_{\mathrm{std}}$ & 0.0114 & 0.0201 \\
\midrule
\multicolumn{3}{l}{\textbf{CIFAR-100} (Swin-Tiny seed-$0$, $b\!=\!2$ for Block-AR)} \\
\midrule
Confidence only & 0.0145 & 0.0191 \\
Predictive entropy & 0.0160 & 0.0156 \\
Aggregate routing entropy $r_{\mathrm{agg}}$ & 0.0151 & 0.0197 \\
Last-layer routing entropy $H_L$ & 0.0157 & 0.0218 \\
Routing concentration $1\!-\!r_{\mathrm{agg}}$ & 0.0151 & 0.0197 \\
Routing entropy $\times$ confidence & 0.0130 & 0.0147 \\
Depth standard deviation $r_{\mathrm{std}}$ & 0.0137 & 0.0212 \\
\bottomrule
\end{tabular}
\end{table}

\subsection{Extended discussion of \S\ref{sec:discussion}}
\label{app:paradigm}

Compressing $(H_1,\ldots,H_L)$ to one scalar can erase information;
Proposition~4 (App.~\ref{app:theory}) separately shows how kernel
smoothing can attenuate a subgroup contrast. These are possible
failure modes, not explanations established by the present data. The
seed-$0$ feature ablation shows no consistent point-estimate advantage
across the tested scalar summaries. On the Swin/CIFAR-10
Block-AR run, $r_{\mathrm{std}}$ has a small but detectable
marginal rank association with the absolute correctness residual
$|\mathrm{conf}(x)-\mathrm{correct}(x)|$ ($\rho=0.0477$,
$p=1.8\times10^{-6}$; Fig.~\ref{fig:mechanism}). That residual is not a
population calibration error, and the statistic does not test conditional
independence given confidence, so no conditional increment follows from it
either way. Cross-seed instability is established
separately for the $r_{\mathrm{agg}}$ diagnostic in
\S\ref{sec:rq2}.

The routing-entropy-profile analysis avoids scalar projection but
still tests a particular learner. It has a small log-loss gain over shuffled
profiles and worse log-loss and Brier than adequate confidence-only prediction
(App.~\ref{app:probe-sanity}). The failed profile sensitivity screen in
App.~\ref{app:profile-sensitivity} prevents reading that latter non-detection
as a tight information bound. Scalar diagnostic or NW sensitivity applies
to its specified projection and teacher; it does not establish sensitivity
to arbitrary entropy-profile interactions.

% ============================================================
\section{Training and evaluation details}
\label{app:details}
All models used in the reported analysis (Swin-Tiny, DeiT-Small, and
ViT-B/16, as named in
Appendix~\ref{app:provenance}) are trained from scratch with the standard DeiT
augmentation recipe: mixup $0.8$, CutMix $1.0$, label smoothing $0.1$, and
RandAugment (\texttt{rand-m9-mstd0.5-inc1}). The objective adds a soft-binned ECE
auxiliary term~\citep{karandikar2021soft} to the cross-entropy, with weight $0.1$,
$15$ soft bins and soft-binning temperature $0.1$, computed on the final logits with
the arg-max of the mixed target as the correctness label; the training loop adds it
to the loss before back-propagation, with and without gradient accumulation. For
all 24 replicated runs, the resolved configuration printed at the start of every
training session enables the term and the per-iteration training log records its
weight throughout; the code revision that ran was not under version control, so its
execution rests on these configurations and logs and on the released code, not on a
separate loss trace. The archived ViT-B/16 AR configurations enable the same term.
AdamW uses a base learning rate of $5\times10^{-4}$ per $512$ examples, scaled
linearly with the effective batch, a cosine-annealed
schedule~\citep{loshchilov2016sgdr} with a $20$-epoch warm-up~\citep{goyal2017accurate}
and mixed precision (AMP at O1)~\citep{micikevicius2017mixed}.

\textbf{Training histories.} According to the training logs, 21 of the 24 runs follow
this recipe in one uninterrupted 300-epoch session with 64 examples per step and two
accumulation steps (effective batch $128$, learning rate $1.25\times10^{-4}$). Three
CIFAR-10 seed-0 runs deviate. DeiT-Small Full-AR used an effective batch of $64$
(learning rate $6.25\times10^{-5}$) in a single 300-epoch session. DeiT-Small Block-AR
and Swin-Tiny Full-AR, the two lower-accuracy runs, started with a 50-epoch cosine
schedule, an effective batch of $64$ and no mixed precision; their logged learning
rate reached its minimum (about $10^{-6}$) near epoch 50 and stayed there after they
were resumed under the 300-epoch configuration at epochs 71 and 101, respectively,
until epoch 299. The Swin-Tiny run used an effective batch of $128$ after resuming.
The released configuration of each run is that of its final session; the supplement
lists every session extracted from the logs. All three runs reached epoch 299 and are
analysed under the fixed inclusion rule; Apps.~\ref{app:effect-size}
and~\ref{app:probe-sanity} report post-hoc analyses without them. Calibration is evaluated on a $50/50$ calibration/test split of the official
test set with fixed seed~$42$. Global equal-width ECE uses $15$ bins; worst-tertile
ECE uses $10$ bins per tertile (App.~\ref{app:wt-ece-illustration}).

\textbf{AR variants.} In every AR stage (the whole network for DeiT-Small and ViT-B/16,
each of the four stages for Swin-Tiny) both the attention and the MLP sub-layer of
each transformer layer attend over a list of earlier states. Full AR appends a state
after every layer in Swin-Tiny and after every sub-layer in DeiT-Small. Block AR
appends the stage input and then one state after every $b$ layers: the Swin-Tiny
Block-AR runs set $b=2$, and the DeiT-Small Block-AR runs use the implementation's
default $b=\max(1,\lfloor 12/8\rfloor)=1$ layer. Omitting sub-layers with a single
available state leaves $L=8$ (Swin Block), $16$ (Swin Full), $22$ (DeiT Block) and
$23$ (DeiT Full) profile entries.

\section{Implementation notes}
\label{app:impl}
Routing weights $\alpha_l$ are extracted by temporarily enabling a cache on
each Attention-Residual sub-layer module; no re-training is required.
The reported AR-CondCal runs use a label-free min--max rescaling of
the routing cache before the fixed 50/50 split. This transform uses
routing values only and no labels; the Nadaraya--Watson fit,
correctness targets, and Scott bandwidth statistics are computed on
$\mathcal{D}_{\mathrm{cal}}$. A cal-only-min/max audit produced
bit-identical Tab.~\ref{tab:calibration_block_ar} values for the
matched-kernel NW rows, so the paper's matched-control conclusions are
unchanged. The main AR-CondCal row uses Scott's-rule bandwidths;
Appendix~\ref{app:bw-sensitivity} reports a bandwidth-sensitivity
sweep over Scott multiples, CV-NLL, and a global-ECE oracle. The AR-CondCal
bisection target $\tilde c(x)$ is projected to
$[1/K\!+\!\epsilon,\, 1\!-\!\epsilon]$ with $\epsilon = 10^{-6}$
before solving for the per-example temperature $\tau(x)$ whose top-class
probability equals the projected target (\S\ref{sec:ar-condcal}); on the
evaluated cells the pooled clipping rate is $0.08\%$ total.

\section{Reproducibility protocol summary}
\label{app:reproducibility}
\textbf{Training.} 300 epochs on Swin-Tiny / DeiT-Small /
ViT-B/$16$, effective batch $128$, AdamW (base lr $5\!\times\!10^{-4}$ per 512
examples, linearly scaled; wd $0.05$), soft-binned ECE auxiliary loss (weight $0.1$),
cosine schedule with $20$-epoch warmup~\citep{loshchilov2016sgdr,goyal2017accurate},
AMP at O1~\citep{micikevicius2017mixed}; three CIFAR-10 seed-0 runs deviate in batch
or schedule (App.~\ref{app:details}).
\textbf{Calibration benchmark.} A $50/50$ cal/test split of the standard
test set with seed $42$; bootstrap CIs on the test half use
$500$ resamples. The diagnostic uses the full test set and
$5{,}000$ within-bin permutations (\S\ref{sec:rq1}).
\textbf{AR-CondCal.} 2-D Nadaraya--Watson on
$(c, r_{\mathrm{std}})$, per-axis Scott's-rule bandwidths on
$\mathcal{D}_{\mathrm{cal}}$ (\S\ref{sec:ar-condcal}); the
kernel-predicted target is projected to
$[1/K\!+\!\epsilon,\, 1\!-\!\epsilon]$ before bisection.
\textbf{Primary entropy-profile probability procedure.} Binary correctness
target, all 24 runs, paired stratified five-fold outer cross-fitting
with two repeats (seeds 100/101). Fold index $k=0,\ldots,9$ uses inner
80/20 split seed $3100+k$ and MLP seed $4100+k$. Profile, zero-padded confidence
and shuffled arms each select between a 32-unit ReLU MLP and the same
regularised logit-confidence logistic reference ($C=1$), using inner-validation
binary log-loss. MLP: Adam, learning rate and weight decay $10^{-3}$,
batch 200, maximum 300 epochs, patience 20, minimum improvement $10^{-5}$;
fresh outer-training refit for the selected epoch count. Scaling is fitted
on inner training for selection and outer training for refitting. Whole
profile rows are permuted separately within inner training/validation and
outer training/test, with seeds $5100+4k+j$, $j=0,1,2,3$. No outer test
labels guide fitting or selection. Score the mean of the two out-of-fold
probabilities using binary log-loss and Brier; AUROC is secondary.
An earlier implementation is described in App.~\ref{app:profile-repair}.
\textbf{Profile uncertainty and sensitivity.} Pooled estimates use 5,000
hierarchical bootstrap draws (seed 0), settings then paired runs, with percentile
intervals; record-weighted and equal-setting pooling coincide in the balanced 24-run
design. The local-$Q$ audit has 20 profile repeats per setting at $0$, $0.005$ and
$0.010$ nats; its stop rule prevented weaker profile strengths and a full sweep
(App.~\ref{app:profile-sensitivity}). The scalar audits cover all five strengths
(Table~\ref{tab:scalar-sensitivity}); they do not validate profile sensitivity.
\textbf{Bandwidth sensitivity.} Multipliers $\{0.25, 0.5, 1, 2,
4\}$; CV-NLL uses $5$-fold cal-set CV; Oracle-ECE selected on
the held-out test half (App.~\ref{app:bw-sensitivity}).

\FloatBarrier
\section{Keeping the reference: a residual correction probe}
\label{app:residual-probe}
Appendix~\ref{app:richer-output} left a specific question open. The augmented
candidate there had to relearn the output mapping, and it lost to the
regularised reference by more than its routing gain could recover. If that
reference is kept and never relearned, can a correction of the same budget turn
routing into a gain over it? This is a follow-up designed after seeing those
results. It is not a preregistered replacement for the original experiment, and
it is not a new algorithmic contribution; conditional probing, zero-padded
baselines and logistic offset models are all established.

\paragraph{Construction.}
Inside each training split we fit the Experiment~2 reference $q_{\rm REF}$
and then learn only a log-odds correction,
\begin{equation}
q_{\rm ARM}=\sigma\!\left(\operatorname{logit}(\operatorname{clip}(q_{\rm REF}(Z)))+h_{\rm ARM}(x_{\rm ARM})\right),
\end{equation}
with $x_{\rm A}=[Z,\mathbf{0}_L]$, $x_{\rm B}=[Z,R]$ and $x_{\rm C}=[Z,\pi(R)]$.
All three arms share one fitted reference, and A receives an output-only
correction on the identical grid and seed budget, so B is never the only arm
given extra capacity. The corrector is the same 32-unit ReLU network, optimiser
and search budget as before; its output layer starts at zero, so each arm
begins exactly at $\operatorname{clip}(q_{\rm REF})$, which is the convention
every model here is scored under. That $h\equiv0$ lies in the function class is
containment, not a finite-sample guarantee of non-inferiority, and each arm
still reverts to the reference when it does not improve on it in inner
validation. Offsets used for training are out-of-fold inside the training
partition under a pre-fixed nested split; offsets used for prediction come from
the reference refit on the data allowed at that stage. Correctors are trained on
sampled binary labels, not on teacher probabilities; teacher probabilities and
oracle gains are used for generation, evaluation and audit, never as training input
or a selection signal. All labels here are simulated; no real correctness label is
used. $\pi$
remains a routing-destruction control, not an exact conditional-null sampler.

\paragraph{What was run, and the continuation rule.}
This is a paired follow-up on simulation records we had already observed, not a
fresh independent validation. Phase~1 covered the two CIFAR-100 settings with
the existing profile teacher and oracle definition unchanged, at $0$, $0.005$
and $0.010$ nats, $20$ replicates per cell: $120$ data replicates and $37{,}200$
model fits. The $120$ data replicates are the simulation units; the fit count is
not a sample size. Seeds are those of the Experiment~2 positive control, and we
verified for all $120$ pairs that the generator manifest hash, the oracle gain,
and the ORACLE and $Q_0$ risks agree, those risks being functions of the drawn
labels and teacher probabilities alone. That is an equality check on derived
quantities, not a byte-level comparison of stored arrays, which the result files
do not retain. Fixing the new code in advance does not turn previously inspected
Experiment~2 results back into unobserved validation data.

One instance, setting~3 at $0.010$ nats replicate~$0$, was executed and
inspected beforehand as a timing and correctness check. It is included in the
summaries and in the replicate counts, and we did not draw replacement seeds.
The Experiment~2 control disclosed the same instance for the same purpose.
Measured cost was $44.6$~s and $243$~MB per replicate on one core. There were no
fit failures, no epoch-cap selections and no optimiser warnings.

A continuation threshold was fixed before running: a positive $95\%$ Monte Carlo
lower bound on the mean routing increment in all four positive-strength
conditions; at least $25\%$ mean recovery of the exact oracle gain at $0.010$
nats; a positive lower bound for the selected routing arm against the reference
at $0.010$; and no unexplained numerical faults. The specification recorded before
the run applies the recovery criterion to the mean across the two settings; a
per-setting check was added afterwards, and both are reported (the specification and
the added check are in the supplementary material). This is an engineering criterion for allocating further effort. It is not a boundary for
the existence of information, not a statistical significance threshold, not a
power certification at a fixed test level, and not a bound on real-data
information.

Table~\ref{tab:residual-probe} in the main text reports every condition run.

\paragraph{Outcome.}
Three criteria were met and one was not. The routing increment has a positive
$95\%$ Monte Carlo lower bound in all four positive-strength conditions
($+0.212$, $+0.656$, $+0.721$ and $+2.105\times10^{-3}$ nats, for means of
$+0.276$, $+0.774$, $+0.814$ and $+2.257$), and the selected routing arm beats
the reference at $0.010$ nats in both settings (selected-arm gain
$G_{\rm sel}=L(\mathrm{REF})-L(B_{\rm sel})$: lower bounds $+0.615$ and $+2.013$, means
$+0.729$ and $+2.190$; Table~\ref{tab:residual-probe} instead lists the unselected
candidate's gain $G_{\rm cand}$, a different predictor). Recovery of the oracle gain at $0.010$
was $8.14\%$ on Swin/CIFAR-100 and $22.57\%$ on DeiT/CIFAR-100, a cross-setting
mean of $15.35\%$. The criterion is missed on the recorded cross-setting mean and on
the added per-setting check. We did not move the threshold, run weaker strengths, add a further
model, or carry the pipeline to the real correctness labels. The result is a
positive but limited recovery under the specified simulation, not an absence of
detected signal.

Under the null the selected correction arm carries a small observed loss
disadvantage relative to the reference ($G_{\rm sel}=-8.8\times10^{-5}$ and
$-5.7\times10^{-5}$ nats, with intervals excluding zero). This is a small disadvantage,
not equivalence; fitting a correction where there is nothing to correct is one
explanation compatible with it, not an isolated cause. The oracle
gain is zero there and no predictor beats the Bayes risk of the known
conditional probability, but since
$\ell_\eta(q)-\ell_\eta(\eta)=D_{\rm KL}(\mathrm{Bern}(\eta)\Vert\mathrm{Bern}(q))\ge 0$
identically, that check verifies the loss definition and the numerics; it is not
a leakage certificate. The leakage boundary rests instead on the nested split,
the training and refit boundaries, and the tests in the artifact, each of which
covers only what it states.

Read at matching predictor identities, the comparison with Experiment~2 is
informative. The earlier \emph{candidate} trailed the reference by $2.7$ to
$7.9\times10^{-3}$ nats and the earlier \emph{selected} contrast was essentially
zero; the residual candidate is level with or ahead of the reference. That
movement is a property of the candidate and includes baseline predictive
improvement, so it is not recovered routing information: the routing quantity is
the same-budget paired increment in the table, which recovers a minority of a
known effect. The teacher here is the existing profile teacher, which already
contains non-linear and interaction structure within the profile; no
output-coordinate by routing teacher was added, and that extension was not run.
Experiment~3's behaviour on the real correctness labels is therefore unknown and
is not evidence of zero increment.

\section{Runs, code and data}
\label{app:provenance}

\begin{table}[!htbp]
\centering\small
\caption{Test accuracy of every run in the 24-run design (full
10,000-example test set). Two lower-accuracy seed-0 runs (Swin-Tiny Full-AR,
0.744, and DeiT-Small Block-AR, 0.695) are retained under the fixed inclusion
rule rather than excluded post hoc by realised accuracy; both reached epoch 299
but did not follow the standard schedule (App.~\ref{app:details}).}
\label{tab:run-accuracy}
\begin{tabular}{llccc}
\toprule
Setting & AR variant & seed 0 & seed 1 & seed 2 \\
\midrule
DeiT-Small / CIFAR-10 & Block-AR & 0.6947 & 0.9055 & 0.9011 \\
DeiT-Small / CIFAR-10 & Full-AR & 0.8972 & 0.9180 & 0.9103 \\
DeiT-Small / CIFAR-100 & Block-AR & 0.6842 & 0.7013 & 0.6880 \\
DeiT-Small / CIFAR-100 & Full-AR & 0.7207 & 0.7205 & 0.7058 \\
Swin-Tiny / CIFAR-10 & Block-AR & 0.9026 & 0.9012 & 0.9024 \\
Swin-Tiny / CIFAR-10 & Full-AR & 0.7444 & 0.9242 & 0.9239 \\
Swin-Tiny / CIFAR-100 & Block-AR & 0.7216 & 0.7264 & 0.7245 \\
Swin-Tiny / CIFAR-100 & Full-AR & 0.7499 & 0.7539 & 0.7580 \\
\bottomrule
\end{tabular}
\end{table}

\textbf{Runs and data integrity.} Table~\ref{tab:run-accuracy} lists the test
accuracy of the 24 replicated runs. Each run has its own epoch-299 checkpoint and
routing cache. The 24 caches have pairwise distinct logits (content hashes are
listed in the supplementary manifest), and the accuracy recomputed from each cache
agrees with the final accuracy in its training log to within $0.0002$. A duplicate
cache file stored under a tag that is not part of the design was found during
auditing and is not used. Two lower-accuracy CIFAR-10 seed-0 runs reached epoch 299
but did not follow the standard schedule (App.~\ref{app:details}); they are retained
under the fixed inclusion rule, and the post-hoc sensitivities of
Apps.~\ref{app:effect-size} and~\ref{app:probe-sanity} omit them, alone or together
with the third deviating run.

\textbf{Training, extraction and evaluation.} Models are trained from scratch with
the code and per-run configuration files included in the supplementary material;
each released configuration is that of the run's final training session, and the
earlier sessions of the deviating runs are documented by a per-session record
extracted from the training logs.
Routing weights are cached by a separate extraction step that reads a checkpoint
and writes, per test example, the logits, the label and the per-layer routing
entropies. The calibration benchmark, the diagnostic and all probes read only these
caches.

\textbf{Supplementary material and reproducibility levels.} The supplementary
archive contains (i) the routing caches of the 24 replicated runs (logits, labels and
entropy profiles); (ii) the code of every analysis in the paper; (iii) per-run results,
pooled summaries and a single manifest that maps each headline number to its source
file and generating command; (iv) the training, extraction and evaluation code with
the run configurations; and (v) an audit record of earlier implementations and
corrections. It does not contain model checkpoints. Three levels of reproduction
should be distinguished. One command re-aggregates the stored per-run results in a
separate copy of the archive: it recomputes the pooled summaries, the 30-test diagnostic
record and Tables 1, 6 and 10--14, redraws Figs.~2--4, and checks every headline value
listed in the supplementary manifest. In our reference environment all recomputed files
were byte-identical to the archived ones and the figures pixel-identical; in other
environments the recomputation is compared with the archive field by field, allowing only
last-digit floating-point differences in derived values. Separate scripts regenerate
Figs.~5--9 from the routing caches, with identical pixels in the reference environment.
Figure~1 is a drawing, Table~3 is hand-curated, and the remaining appendix tables are
assembled from stored per-run outputs listed in the supplementary manifest. Refitting
from the included caches is possible for every analysis of the 24 replicated runs (the
six ViT-B/16 diagnostic values are archived without caches): the archive's smoke
test refits the diagnostic, the continuous-confidence check, both probes and the
calibration benchmark on one run and reproduces the stored values (to $10^{-15}$; to
$1.2\times10^{-7}$ for the benchmark, whose stored values were produced with earlier
library versions). A refit of all runs is estimated at two to three hours on one
multi-core node. Re-extracting caches or retraining models requires the checkpoints
or training runs and cannot be verified from the archive alone.

\end{document}